\documentclass[lettersize,journal]{IEEEtran}
\usepackage{amsmath,amsfonts}
\usepackage{algorithmic}
\usepackage{algorithm}
\usepackage{array}
\usepackage[caption=false,font=normalsize,labelfont=sf,textfont=sf]{subfig}
\usepackage{textcomp}
\usepackage{stfloats}
\usepackage{multirow}
\usepackage{booktabs}
\usepackage{url}
\usepackage{verbatim}
\usepackage{graphicx}
\usepackage{subcaption}
\usepackage{cite}
\usepackage[numbers, sort&compress]{natbib}
\begin{document}

\title{End-to-end Autonomous Vehicle Following System using Monocular Fisheye Camera}

\author{Jiale Zhang, Yeqiang Qian\textsuperscript{*}, Tong Qin, Mingyang Jiang, Siyuan Chen, and Ming Yang\textsuperscript{*}
  \thanks{Jiale Zhang, Yeqiang Qian, Mingyang Jiang, Siyuan Chen and Ming Yang are with the department of Automation, Shanghai Jiao Tong University, Shanghai, 200240; Key Laboratory of System Control and Information Processing, Ministry of Education, Shanghai, 200240, China (mingyang@sjtu.edu.cn). Corresponding authors: Yeqiang Qian; Ming Yang.}
  \thanks{Tong Qin is with the Global Institute of Future Technology, Shanghai Jiao Tong University, Shanghai, 200240, China.}
  \thanks{This work is supported in part by the National Natural Science Foundation of China under Grants 62173228.}

}



\maketitle

\begin{abstract}

The increase in vehicle ownership has led to increased traffic congestion, more accidents, and higher carbon emissions. Vehicle platooning is a promising solution to address these issues by improving road capacity and reducing fuel consumption. However, existing platooning systems face challenges such as reliance on lane markings and expensive high-precision sensors, which limits their general applicability. To address these issues, we propose a vehicle following framework that expands its capability from restricted scenarios to general scenario applications using only a camera. This is achieved through our newly proposed end-to-end method, which improves overall driving performance. The method incorporates a semantic mask to address causal confusion in multi-frame data fusion. Additionally, we introduce a dynamic sampling mechanism to precisely track the trajectories of preceding vehicles. Extensive closed-loop validation in real-world vehicle experiments demonstrates the system’s ability to follow vehicles in various scenarios, outperforming traditional multi-stage algorithms. This makes it a promising solution for cost-effective autonomous vehicle platooning. A complete real-world vehicle experiment is available at https://youtu.be/zL1bcVb9kqQ.

\end{abstract}


\section{Introduction}

\IEEEPARstart{T}{he} rapid growth of global urbanization has driven a continuous increase in the demand for freight transportation, resulting in a significant increase in the number of motor vehicles on the roads and the distances they travel \cite{bayliss2009motoring}.
To address transportation challenges, the concept of vehicle platooning has been introduced. Vehicle platooning refers to the closely coordinated movement of vehicles within a safe distance \cite{maiti2017conceptualization}. It holds promise for improving road capacity through reduced inter-vehicle distances and lowering fuel consumption, with plans to expand its application to public transportation \cite{9852980,7970188,dey2015review}.

Mainstream vehicle following systems are designed primarily for structured roads, such as highways. These systems rely on Cooperative Adaptive Cruise Control (CACC) \cite{8598907} for longitudinal control and lane marking detection for lateral control \cite{10106476}. Although these technologies enable effective vehicle following in structured environments, their reliance on lane markings can lead to challenges in close formations, where occlusions caused by preceding vehicles make lane detection unreliable \cite{solyom2013performance}. Consequently, their application is generally restricted to highways, limiting their adaptability to general scenarios.

Some systems have also attempted to address vehicle following in general scenarios by tracking the historical trajectory of the preceding vehicle, eliminating the reliance on lane markings. However, achieving this requires precise perception of the longitudinal and lateral positions of the preceding vehicle.
To meet these requirements, these systems are dependent on expensive sensors, including real-time kinetic global positioning system (RTK GPS) equipment for centimeter-level relative localization, LiDAR for high-precision detection, and V2V communication devices for data exchange between vehicles \cite{1570755, 8598907, yi2023lidar}, which increases the complexity and cost of the system, making widespread adoption impractical. Existing vehicle following systems face critical challenges, either being restricted by structured roads or requiring expensive sensors, which limits their practicality. To address these challenges, our research focuses on developing a vehicle following system that can be applied to general scenarios using a single monocular camera.

A key limitation of traditional vehicle following systems is their reliance on multi-stage methods. These methods require explicit transmission of intermediate results, which often leads to cumulative perception errors \cite{10258330}. To address these challenges, we propose an end-to-end vehicle following framework. By integrating multiple modules through joint training, our approach unifies the optimization objectives across the entire model and minimizes error accumulation between modules. This reduces perception errors introduced by low-cost sensors, allowing robust vehicle following in general scenarios with just a single camera. However, existing end-to-end algorithms face two major challenges in the application of vehicle following tasks. First, these algorithms are susceptible to causal confusion \cite{chen2024endtoendautonomousdrivingchallenges}, which restricts their applicability to open-loop testing on datasets or validation in simulation environments, preventing their deployment on real-world vehicles. Second, accurately estimating the historical trajectory and motion information of the preceding vehicle using only image data is challenging. 

To mitigate the causal confusion phenomenon during multi-frame fusion, we analyze its causes and resolve it by filtering specific features using semantic masks. This approach not only eliminates causal confusion, but also improves the computational efficiency of the network. Additionally, traditional methods that sample multiple frames at fixed time intervals fail to provide sufficient historical trajectory information of the preceding vehicle. To overcome this limitation, we propose a dynamic sampling mechanism combined with a multi-frame feature fusion module. This method dynamically adjusts the sampling process based on changes in the spatial distance of the preceding vehicle, allowing more accurate inference of its historical trajectory and effectively avoiding potential issues related to the status of the ego vehicle in end-to-end systems \cite{li2024egostatusneedopenloop}. To our knowledge, we are the first to achieve general scenario autonomous vehicle following using only a single camera and to apply an end-to-end algorithm in real-world vehicle following tasks. Overall, the main contributions of this paper are as follows.

\begin{itemize}
  \item We designed an end-to-end method that uses only a camera to complete the vehicle following task in general road environments. We accomplished a more complex task with fewer sensors, reducing the cost of vehicle following in various scenarios.
  \item We introduced a semantic mask to eliminate causal confusion during multi-frame fusion in Bird’s Eye View (BEV) perception networks. It filters out irrelevant features and enables the neural network to learn correct causal reasoning. We also proposed a dynamic sampling mechanism to tackle the challenge of learning the preceding vehicle’s historical trajectory in the vehicle following task. It samples frames by spatial distance for multi-frame fusion and provides the neural network with richer spatio-temporal information.
  \item We conducted closed-loop validation of the algorithm in real-world vehicles. The results demonstrate that the algorithm can stably follow the preceding vehicle in various scenarios. Our method outperforms existing multi-stage algorithms in vehicle following performance, showcasing its effectiveness and reliability in practical applications.
\end{itemize}

\section{Related Work}
\subsection{Vehicle following systems}

The earliest research on vehicle following systems dates back to the California Partners for Advanced Transit and Highways (PATH) project in the late 1980s \cite{69979}.
This pioneering work inspired subsequent research and real-world demonstrations around the world \cite{fritz1999longitudinal,bergenhem2010challenges,bergenhem2012vehicle}.
Building on this foundational research, current vehicle following systems rely on a multi-stage method to accomplish the task.
Typically, onboard sensors such as cameras, LiDAR, or radar are used to perceive the position and velocity of the preceding vehicle.
This explicit perception result is then passed downstream to a controller to follow the preceding vehicle \cite{8569947}. Tang et al. utilized a vision-based system that integrates the YOLO v3 algorithm for object detection and distance measurement from driving images \cite{9687583}. Hsu et al. incorporated LiDAR for environment detection and relative position tracking, supplemented by a camera for verification and data fusion \cite{6425157}. These methods are used primarily to detect the longitudinal distance to the preceding vehicle, which is only applicable on structured roads, such as highways with clear lane markings. However, a vehicle following system that tracks the historical trajectory of the preceding vehicle must simultaneously detect both the lateral and longitudinal positions. In \cite{1570755}, the vehicle following problem was converted into a path-following problem using high-precision RTK GPS sensors and inter-vehicle communication for trajectory reference, achieving high tracking accuracy. However, the high cost of RTK GPS equipment and potential signal loss in urban environments limit its feasibility for mass deployment.

In terms of control, vehicle following task is divided into longitudinal and lateral control. For longitudinal control, CACC is commonly used to maintain safe distances. Current research on CACC focuses mainly on the design of distributed controllers \cite{6427034,wang2014control,di2014distributed}, improvements to Model Predictive Controllers \cite{kianfar2012design,zheng2016distributed}, and addressing communication delays \cite{di2014distributed,di2015design,jia2016platoon}. As for lateral control, vehicle following systems that track the historical trajectory of the preceding vehicle are responsible for completing the trajectory tracking task, with research focusing primarily on improving control accuracy \cite{1998,6722531,7535437}. Overall, each module in the multi-stage vehicle following method has different optimization objectives. When the upstream perception module has larger errors, it becomes difficult for the downstream control module to correct these errors, leading to error propagation and affecting the optimization of the entire system's performance.

\subsection{End-to-end algorithms for autonomous driving}

End-to-end algorithms have gained widespread attention for their ability to eliminate errors between different modules \cite{chen2024endtoendautonomousdrivingchallenges}. These algorithms can be categorized into reinforcement learning (RL)-based and imitation learning-based approaches. Most research on end-to-end autonomous driving based on RL has been restricted to simulations, such as the CARLA simulator \cite{Dosovitskiy17}, due to the need for extensive data and safe training environments, which are difficult to achieve in the real world. Chen et al. developed an RL-based vision policy using a `world on rails' assumption to compute action-values via dynamic programming \cite{9709920}. Peng et al. proposed an expert-guided policy optimization method that incorporates a guardian into reinforcement learning to improve safe exploration and policy optimization \cite{peng2021safedrivingexpertguided}. Li et al. proposed HACO, a human-in-the-loop RL method that uses human interventions to train safe and efficient autonomous driving policies without environmental rewards \cite{li2022efficientlearningsafedriving}. 
Despite these advancements, the transfer of the simulation results to real-world scenarios remains a major challenge.

Imitation learning primarily utilizes the Behavior Cloning method, which formulates the task as supervised learning to minimize the gap between the agent's and expert's actions. 
Prakash et al. introduced TransFuser \cite{prakash2021multimodalfusiontransformerendtoend}, a Multi-Modal Fusion Transformer that combines images and point clouds with attention mechanisms, excelling in complex urban driving. Chen et al. improved driving policy training by learning from all observed vehicles, increasing scenario diversity and prediction precision without additional data collection \cite{chen2022learningvehicles}. Zhang et al. developed Roach, which supervises imitation learning agents to achieve expert-level performance in urban driving \cite{zhang2021endtoendurbandrivingimitating}. Hu et al. introduced UniAD \cite{hu2023planningorientedautonomousdriving}, a unified autonomous driving framework that integrates perception, prediction, and planning tasks into a single network, optimizing them in a planning-oriented manner. Despite its effectiveness, imitation learning faces issues such as causal confusion, where models learn misleading correlations, as noted in \cite{wen2020fightingcopycatagentsbehavioral,wen2021keyframefocusedvisualimitationlearning}. This often results in excellent performance during open-loop validation, but poor results in closed-loop testing. In addition, many studies rely on simulation validation rather than real-world driving conditions.

\section{Methods}

\subsection{Problem formulation}

In the context of vehicle following tasks, we employ the principle of Behavior Cloning, which entails training a model by imitating the actions of an expert. Behavior Cloning treats this task as a supervised learning problem, aiming to minimize the discrepancy between the model's outputs and the expert's actions. Specifically, given a dataset \( \mathcal{D} \) of state-action pairs \( (s, a) \), the objective is to minimize the loss between the model’s output \( \pi(s) \) and the expert-provided action \( a \):
\begin{equation}
  \min_{\theta} \mathbb{E}_{(s,a) \sim \mathcal{D}} \left[ \mathcal{L}(\pi(s; \theta), a) \right]  
\end{equation}
where \( \mathcal{L} \) represents the loss function, and \( \theta \) are the model parameters. For the vehicle following task, we extend this framework to primarily infer the planned trajectory of the ego vehicle, as well as the multi-frame pose and velocity of the preceding vehicle. Given a sequence of front-facing fisheye camera images \( \{I_t\}_{t=1}^T \), the goal is to estimate: (1) the planned trajectory of the ego vehicle, consisting of waypoints \( \mathbf{W}_T = \{(x_T^k, y_T^k)\}_{k=1}^{N} \), where \( N \) is the number of waypoints; (2) the pose of the preceding vehicle in BEV space \( \{\mathbf{P}_t\}_{t=1}^{T} \); and (3) the velocity of the preceding vehicle at the current frame \( v_T \). Here, \( \mathbf{P}_t = (x_t, y_t, \varphi_t) \) denotes the position \( (x_t, y_t) \) and orientation \( \varphi_t \) of the preceding vehicle at time \( t \). The loss function for these estimations is formulated as:
\begin{equation}
  \begin{aligned}
    \mathcal{L}(\theta) = & \underbrace{\left\|\pi_w(I_{1:T}; \theta) - \mathbf{W}_T^*\right\|^2}_{\mathcal{L}_w(\theta)} + \underbrace{\frac{1}{T} \sum_{t=1}^T \left\|\pi_p(I_{1:T}; \theta) - \mathbf{P}_{t}^*\right\|^2}_{\mathcal{L}_p(\theta)} \\
                          & + \underbrace{\left\|\pi_v(I_{1:T}; \theta) - v_{T}^*\right\|^2}_{\mathcal{L}_v(\theta)}
  \end{aligned}
\end{equation}
Here, \( \pi_w(I_{1:T}; \theta) \), \( \pi_p(I_{1:T}; \theta) \), and \( \pi_v(I_{1:T}; \theta) \) represent the predicted trajectory waypoints, pose, and velocity based on the entire input sequence \( I_{1:T} \), while \( \mathbf{W}_T^* \), \( \mathbf{P}_{t}^* \), and \( v_T^* \) denote the expert ground truth values.

\begin{figure}[t]
  \centering
  \includegraphics[width=0.5\textwidth]{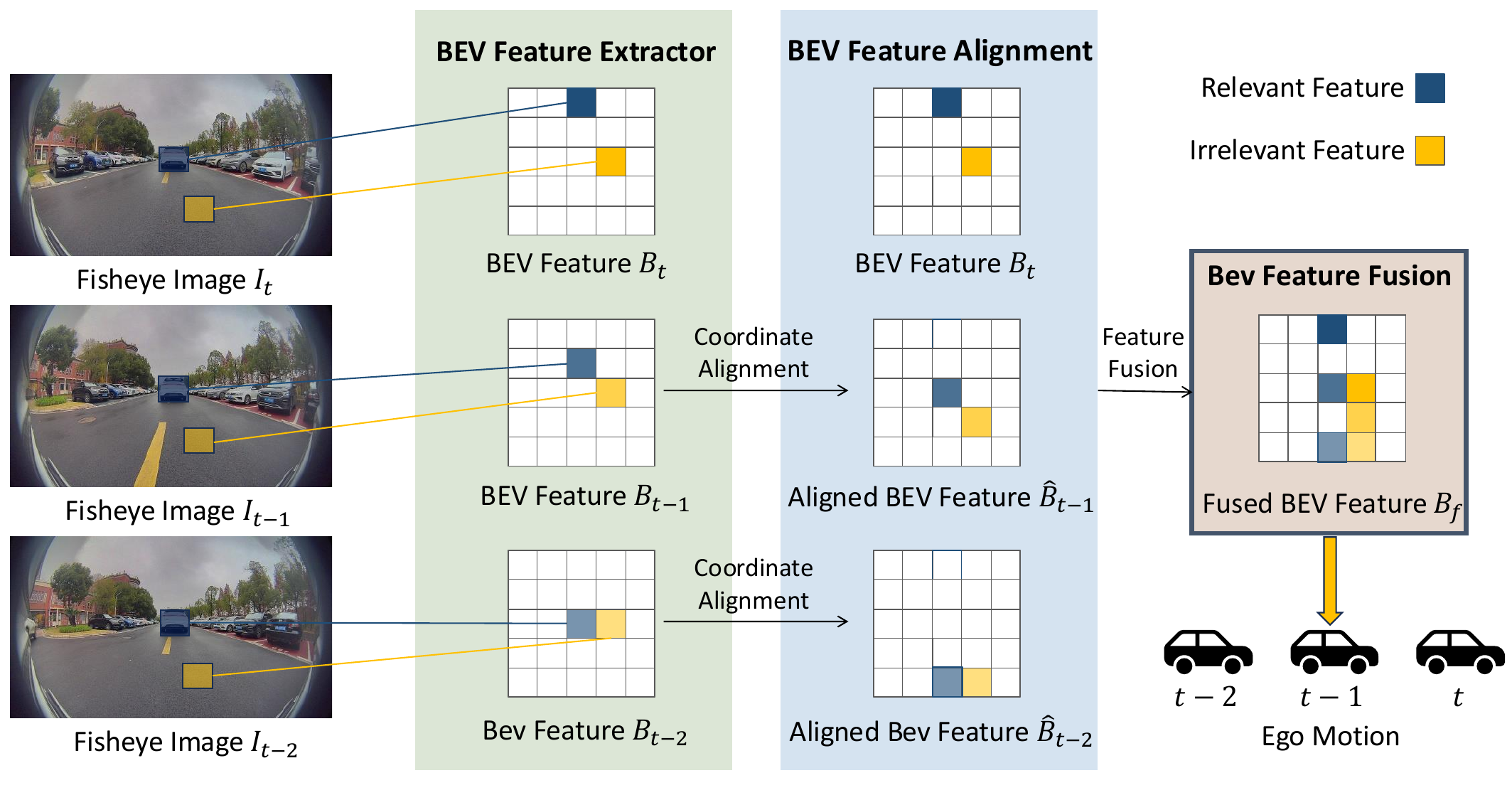}
  \caption{Causal confusion in BEV temporal fusion, where irrelevant features in BEV space shift corresponding to the ego vehicle's past motion.}
  \label{fig:casual}
    \vspace{-15pt}
\end{figure}

\subsection{System overview}

\begin{figure*}
  \centering
  \includegraphics[width=\textwidth]{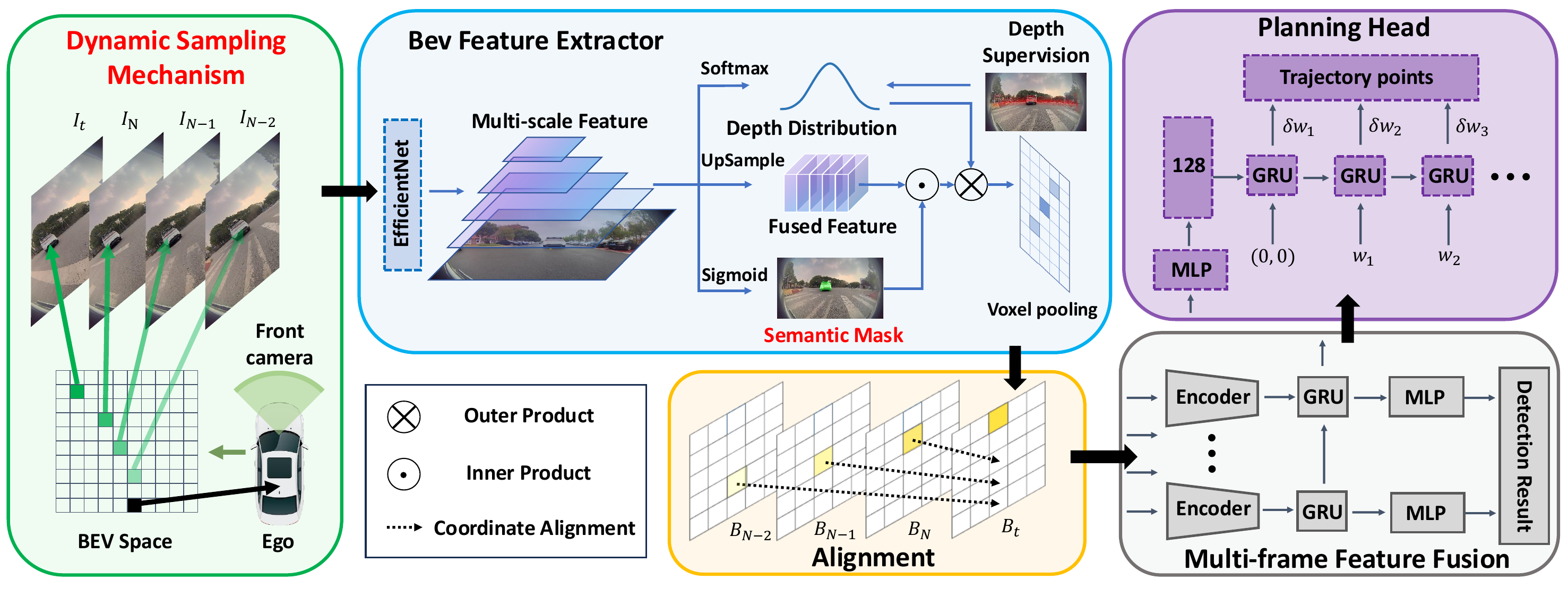}
  \caption{The overall framework of the system. The network processes multi-frame fisheye images to output the planned trajectory of the ego vehicle.}
  \label{fig:system}
  \vspace{-15pt}
\end{figure*}
The overall architecture of the system is illustrated in Figure \ref{fig:system}. The network takes multi-frame fisheye images as input and outputs the planned trajectory for the ego vehicle. First, the dynamic sampling mechanism is used to select multiple fisheye images from the image queue based on appropriate spatial distances. For each input image, EfficientNet is used to extract multi-scale features. The depth distribution and semantic mask for each grid in the extracted feature map are then explicitly estimated. The semantic mask performs a binary classification task to distinguish between the grids corresponding to the preceding vehicle and the other grids. Then it filters out the other grids, projecting only the relevant features into the BEV space based on the depth distribution. The BEV features for each frame are then extracted, and the features from past frames are aligned with the current frame's coordinate system. These multi-frame features are fused using a Gated Recurrent Unit (GRU) encoder, supervised by the detection results of the preceding vehicle. Finally, the fused features are input into the planning head to output the trajectory of the ego vehicle for the next three seconds.

\subsection{Semantic mask to solve causal confusion}

Driving is a task with temporal continuity, where each action is typically closely related to the previous one, allowing future behaviors to be predicted based on past motions. However, when multi-frame data are used during model training, this temporal continuity can lead the model to rely too heavily on past information. For example, this can cause the model to degrade into maintaining constant velocity movement based on the current velocity, preventing it from responding effectively to changes in the preceding vehicle's behavior \cite{li2024egostatusneedopenloop}. This issue is a form of causal confusion, where reliance on historical data results in a decrease in model performance during deployment \cite{geirhos2020shortcut,de2019causal}.
LeCun et al. were the first to observe and report this effect, proposing the use of a single frame input for steering prediction to avoid the issue of causal confusion \cite{muller2005off}. Although this method is relatively simple, it remains the preferred solution in current state-of-the-art imitation learning methods \cite{chen2024endtoendautonomousdrivingchallenges}. However, since single-frame input essentially provides no temporal information, such as the velocity of surrounding vehicles, it fails to allow the network to both receive this temporal data and learn the correct relationships from it.

In the vehicle following task, causal confusion arises during the alignment of multi-frame BEV features. The `Principle of Least Effort' of neural networks encourages the model to rely on shortcut learning \cite{geirhos2020shortcut}, which leads to the network learning incorrect causal relationships. For an input sequence of images \( \{I_t, I_{t-1}, I_{t-2}\} \), the BEV network generates BEV features by projecting image features into the BEV space, denoted as \( \{B_t, B_{t-1}, B_{t-2}\} \). However, not all BEV features contribute equally to the task. These features can be broadly categorized into relevant features and irrelevant features. Relevant features \( B_{\text{rel}} \), derived from the pixels of the preceding vehicle, encode critical information about the vehicle's pose and facilitate the convergence of loss functions. In contrast, irrelevant features \( B_{\text{irr}} \), originate from elements such as roadside pixels and lack task-relevant information, thus providing no meaningful contribution to the task.

We observed that the network generates similar features at the same BEV spatial locations across different time steps, making it less sensitive to the input images. The BEV features of the past two frames, \( B_{t-1} \) and \( B_{t-2} \), are aligned with the current frame's BEV coordinate system \( B_t \). At each time step, the BEV coordinate system corresponds to the ego vehicle's coordinate system at that time. The alignment of features from \( B_{t-1} \) and \( B_{t-2} \) to the current frame \( B_t \) is achieved using rotation matrices \( R_{t-1}, R_{t-2} \) and translation matrices \( T_{t-1}, T_{t-2} \), which represent the transformation from the coordinate system of the ego vehicle at time \( t-1 \) and \( t-2 \) to the coordinate system at time \( t \). The final fused feature \( B_f \), which includes the current frame's BEV feature \( B_t \) and the transformed features \( \hat{B}_{t-1} \) and \( \hat{B}_{t-2} \), can be illustrated as:
\begin{equation}
  \begin{aligned}
    \hat{B}_{t-i} & = R_{t-i} B_{t-i} + T_{t-i}, \quad i = 1, 2, \ldots, n \\
    B_f           & = B_t + \sum_{i=1}^{n} \hat{B}_{t-i}
  \end{aligned}
\end{equation}

this process causes the positional changes of irrelevant features \( B_{\text{irr}} \) in the BEV space to correspond to the past motion of the ego vehicle over the past two frames, as shown in Figure \ref{fig:casual}. The network extracts information not directly from the irrelevant features themselves but from their changes in spatial position, denoted as \( \Delta_{\text{pos}}(B_{\text{irr}}) \), leading to past motion leakage.

Ideally, the network should only use relevant features \( B_{\text{rel}} \) to optimize loss of trajectory planning \( L_w \). However, due to the shortcut behavior, the network often relies on changes in spatial position of irrelevant features \( \Delta_{\text{pos}}(B_{\text{irr}}) \), leading to a suboptimal solution. The total gradient of \( L_w \) with respect to the network parameters \( \theta \) can be expressed as the sum of two terms: one for the relevant features \( B_{\text{rel}} \) and one for the positional changes of the irrelevant features \( \Delta_{\text{pos}}(B_{\text{irr}}) \). This total gradient is given by:
\begin{equation}
\begin{aligned}
      \nabla_{\theta} L_w(\theta; B_{\text{rel}}, \Delta_{\text{pos}}(B_{\text{irr}})) = & \frac{\partial L_w(\theta)}{\partial B_{\text{rel}}} \cdot \nabla_{\theta} B_{\text{rel}} + \\
  & \frac{\partial L_w(\theta)}{\partial \Delta_{\text{pos}}(B_{\text{irr}})} \cdot \nabla_{\theta} \Delta_{\text{pos}}(B_{\text{irr}})
\end{aligned}
\end{equation}

In an optimal scenario, the term \( \frac{\partial L_w(\theta)}{\partial \Delta_{\text{pos}}(B_{\text{irr}})} \) should be zero, meaning the network parameters should not rely on irrelevant feature updates. However, in practice, this term is often larger than the gradient of relevant features $\frac{\partial L_w(\theta)}{\partial B_{\text{rel}}}$, causing the network to update its parameters in a direction dominated by these large gradients, even though the direction is incorrect. This leads the network to converge towards a policy \( \pi_w \) that completely misunderstands causal relationships, making it impossible to effectively deploy in a closed-loop system. To address this issue, we introduced a semantic mask in the BEV space that blocks gradients from irrelevant features, as described in Section \ref{3.4}.

\subsection{BEV feature extraction}
\label{3.4}
This section explains the process of extracting image features and projecting them into the BEV space, starting with fisheye image inputs. The input is a fisheye image \(\mathbf{I}\), represented as \(\mathbf{I} \in \mathbb{R}^{3 \times H \times W}\). To preserve the wide field of view, we extract 2D features directly from the fisheye image without performing distortion correction. First, the image is passed through EfficientNet to extract multi-scale features. Four scales of features are obtained: \(\mathbf{F}_1 \in \mathbb{R}^{C_1 \times \frac{H}{4} \times \frac{W}{4}}\), \(\mathbf{F}_2 \in \mathbb{R}^{C_2 \times \frac{H}{8} \times \frac{W}{8}}\), \(\mathbf{F}_3 \in \mathbb{R}^{C_3 \times \frac{H}{16} \times \frac{W}{16}}\), and \(\mathbf{F}_4 \in \mathbb{R}^{C_4 \times \frac{H}{32} \times \frac{W}{32}}\). Next, all feature maps are upsampled to the size of \(\frac{H}{4} \times \frac{W}{4}\). These multi-scale features are then fused using multiple convolutional layers to generate the final feature map \(\mathbf{F} \in \mathbb{R}^{C \times \frac{H}{4} \times \frac{W}{4}}\):
\begin{align}
   & \mathbf{F}_i = \text{EfficientNet}(\mathbf{I}), \quad i = 1, 2, 3, 4 \\
   & \mathbf{F} = \text{Conv} \left( \mathbf{F}_1, \text{up}(\mathbf{F}_2), \text{up}(\mathbf{F}_3), \text{up}(\mathbf{F}_4) \right)
\end{align}
To estimate the depth distribution at each grid location of the feature map \(\mathbf{F}\), we discretize the depth range into \(N\) depth values \(\{d_1, d_2, \ldots, d_N\}\) with a step size \(\Delta d\), where \(d_i = d_{\min} + (i-1)\Delta d\). The depth network applies a convolutional layer and a softmax activation function to the feature map \(\mathbf{F}\), producing the depth probability distribution \(\mathbf{D} \in \mathbb{R}^{N \times \frac{H}{4} \times \frac{W}{4}}\) for each grid location \((u, v)\). Specifically, for each location \((u, v)\), the depth distribution is:
\begin{equation}
  \mathbf{D}(u, v) = \left[ \mathbf{P}(d_i \mid \mathbf{F}(u, v)) \right]_{i=1}^{N}
\end{equation}
where \((u, v) \in \left\{1, \ldots, \frac{H}{4}\right\} \times \left\{1, \ldots, \frac{W}{4}\right\}\). In this way, the network provides a probability distribution over discrete depth values for each spatial location in the feature map.

\begin{figure}[t]
  \centering
  \includegraphics[width=0.5\textwidth]{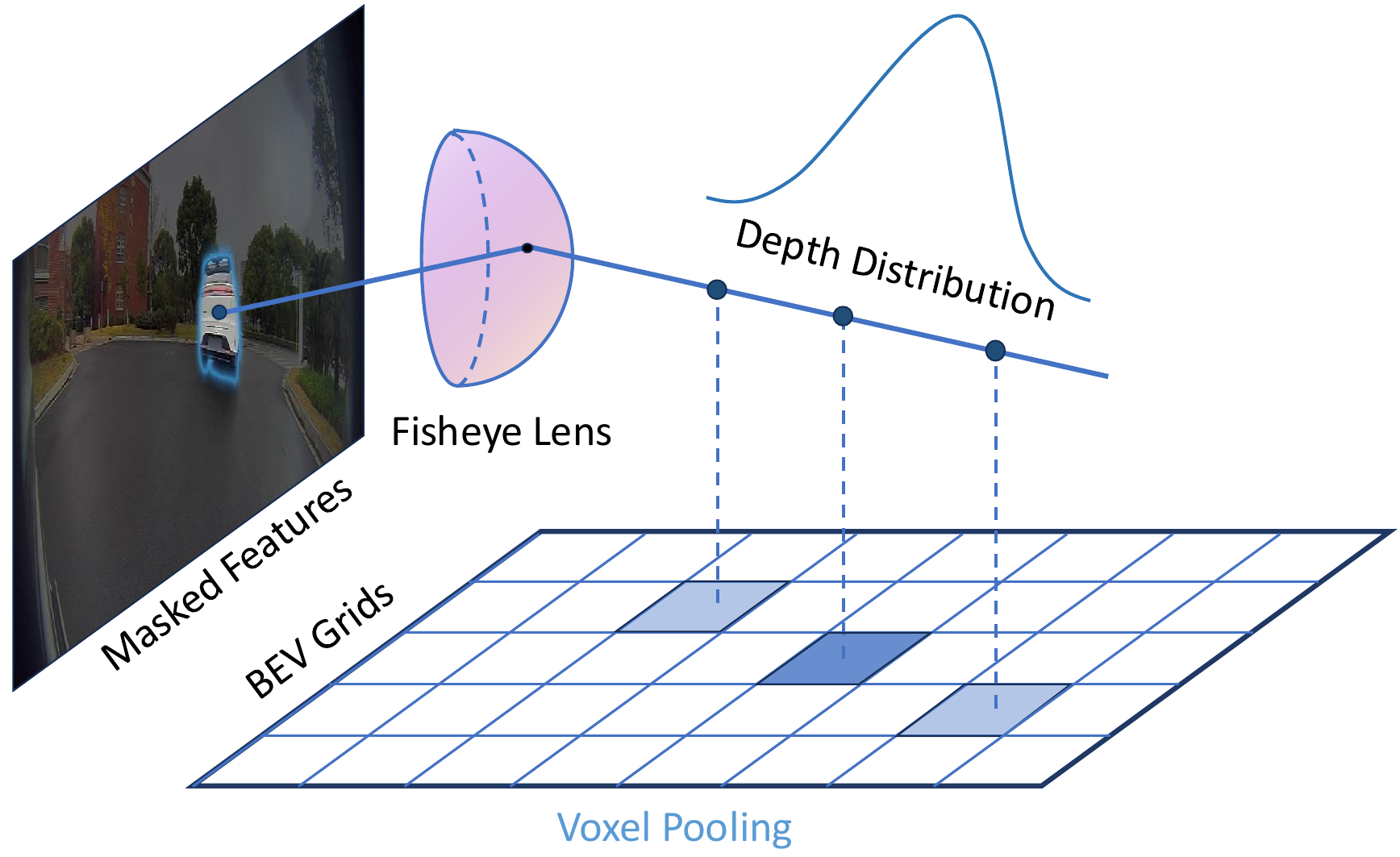}
  \caption{BEV feature extraction with semantic mask. Only relevant features are projected into the BEV space.}
  \label{fig:feature}
    \vspace{-15pt}
\end{figure}

Simultaneously, a semantic mask \( S \in \mathbb{R}^{ \frac{H}{4} \times \frac{W}{4}} \) is estimated for the feature map using a convolutional layer followed by a sigmoid activation function, where each element indicates the probability of the corresponding grid belonging to the preceding vehicle. This distinguishes grids representing the preceding vehicle from others. Then, the semantic mask is binarized based on a predefined threshold \( \tau \) to filter out irrelevant features:
\begin{equation}
  S = \sigma(\text{Conv}(\mathbf{F}))
\end{equation}
\begin{equation}
  \hat{S} = \begin{cases}
    1, & \text{if } S \geq \tau \\
    0, & \text{otherwise}
  \end{cases}
\end{equation}
\begin{equation}
  \tilde{F} = \hat{S} \odot \mathbf{F}
\end{equation}
Here, \( \sigma \) denotes the sigmoid activation function, while \( \odot \) represents element-wise multiplication, and \( \hat{S} \) is the binarized semantic mask obtained by applying the threshold \( \tau \) to \( S \). The filtered features \( \tilde{F} \) primarily retain relevant information related to the preceding vehicle, as shown in Figure \ref{fig:feature}. By applying this semantic mask, the network projects only the relevant features into the BEV space, effectively blocking the influence of the positional changes of irrelevant features \( \Delta_{\text{pos}}(B_{\text{irr}}) \) in the BEV space. This reduces the impact of large gradients from irrelevant features \( \frac{\partial L_w(\theta)}{\partial \Delta_{\text{pos}}(B_{\text{irr}})} \) and forces the network to update its parameters along the correct gradient direction \( \frac{\partial L_w(\theta)}{\partial B_{\text{rel}}} \), addressing the issue of causal confusion.

After obtaining the filtered features \(\tilde{F}\) and the depth distribution \(\mathbf{D}\), the next step is to project these features into the BEV space. To achieve this, the distorted coordinates \((u, v)\) are first converted into ideal coordinates \((u', v')\) using a stretching matrix to correct the fisheye image distortion:
\begin{equation}
  \begin{pmatrix}
    u \\
    v
  \end{pmatrix}
  =
  \begin{pmatrix}
    c & d \\
    e & 1
  \end{pmatrix}
  \begin{pmatrix}
    u' \\
    v'
  \end{pmatrix}
  +
  \begin{pmatrix}
    c_x \\
    c_y
  \end{pmatrix}
\end{equation}
where \(c, d, e\) are the matrix coefficients that account for distortion, and \((c_x, c_y)\) represents the distortion center offset. Next, these ideal coordinates \((u', v')\) are transformed into 3D camera coordinates \((X_c, Y_c, \mathbf{D}(u, v))\):
\begin{equation}
  \begin{pmatrix}
    X_c \\
    Y_c \\
    \mathbf{D}(u, v)
  \end{pmatrix}
  =
  \lambda
  \begin{pmatrix}
    u' \\
    v' \\
    a_0 + a_2 \rho^2 + a_3 \rho^3 + a_4 \rho^4
  \end{pmatrix}
\end{equation}
where \(\rho = \sqrt{u'^2 + v'^2}\) is the radial distance, and \(a_0, a_2, a_3, a_4\) are the Scaramuzza model \cite{4059340} coefficients for radial distortion correction. Finally, the filtered features \(\tilde{F}\) and depth distribution \(D\) are combined via an outer product and projected into the BEV space through voxel pooling based on the constructed frustum. This transformation follows the method used in the LSS algorithm \cite{philion2020lift}, resulting in the final BEV feature representation \(B\):
\begin{equation}
  B = \text{VoxelPooling}\left(\tilde{F} \otimes \mathbf{D}, \text{Frustum}(u, v, \mathbf{D}(u, v))\right)
\end{equation}
\subsection{Temporal fusion}

\subsubsection{Dynamic sampling mechanism}

When the preceding vehicle is stationary or moving at low speed, sampling frames at fixed time intervals results in high similarity between consecutive frames and very small spatial distances between them. This makes it difficult to provide sufficient historical trajectory information, as the frames contain nearly identical spatial data. To overcome this limitation and enable the network to better learn the historical trajectory of the preceding vehicle, we propose a dynamic sampling mechanism that uniformly samples based on spatial distance. This approach allows the network to extract sufficient historical trajectory information, even when the preceding vehicle is stationary or moving slowly. The proposed algorithm for dynamic sampling based on spatial distance is detailed in Algorithm \ref{alg:dynamic_spatial_sampling}. It maintains an image queue to store frames based on the spatial distance traveled by the preceding vehicle. During inference, the preceding vehicle’s pose corresponding to each frame is roughly estimated by the network. Frames are sampled dynamically using a computed spatial interval. If the spatial distance between the ego vehicle and the preceding vehicle becomes smaller than the accumulated distance of frames in the queue, older frames are removed to keep the queue relevant and efficient.
\begin{algorithm}
  \caption{Dynamic Spatial-Temporal Sampling with Queue Management}
  \label{alg:dynamic_spatial_sampling}
  \begin{algorithmic}[1]
    \STATE \textbf{Inputs}: Spatial threshold $\sigma \in \mathbb{R}^+$, Number of historical frames to sample $N \in \mathbb{N}$, Current ego-vehicle pose $p_{\text{ego}} \in \mathbb{R}^n$, Preceding vehicle pose sequence $\mathcal{P} = \{p_k\}_{k=1}^\infty$, Image sequence $\mathcal{I} = \{I_k\}_{k=1}^\infty$, where $p_k$ corresponds to the preceding vehicle pose in $I_k$
    \STATE \textbf{Initialization}: Dynamic frame queue $\mathcal{Q} \leftarrow \emptyset$, Cumulative distance $\Delta S \leftarrow 0$

    \FOR{each incoming image $I_k \in \mathcal{I}$}
      \STATE Compute inter-frame distance of preceding vehicle: $S_k = ||p_k - p_{k-1}||_2$
      \STATE Update cumulative distance: $\Delta S \leftarrow \Delta S + S_k$
      
      \IF{$S_k \geq \sigma$}
        \STATE Enqueue frame with timestamp: $\mathcal{Q}.\text{enqueue}((I_k, t_k))$
        \STATE Reset local distance: $\Delta S \leftarrow 0$
      \ENDIF

      \STATE Compute distance from preceding vehicle to ego-vehicle: $D_k = ||p_{\text{ego}} - p_k||_2$, and compute queue spatial span: $\Lambda = |\mathcal{Q}| \cdot \sigma$

      \IF{$|\mathcal{Q}| \geq N$ and $D_k \leq \Lambda$}
        \STATE Compute sampling interval: $\tau = \frac{D_k}{N + 1}$, define fractional index step: $\eta = \lfloor \frac{\tau}{\sigma} \rfloor$, initialize sampled subset: $\mathcal{Q}_s \leftarrow \emptyset$
        \FOR{$i = 0$ \TO $N-1$}
          \STATE Select frame index: $j = \lfloor i \cdot \eta \rfloor$
          \STATE $\mathcal{Q}_s \leftarrow \mathcal{Q}_s \cup \{\mathcal{Q}[j]\}$
        \ENDFOR
      \ENDIF

      \WHILE{$\Lambda > D_k$}
        \STATE Dequeue oldest frame: $(I_{\text{old}}, t_{\text{old}}) \leftarrow \mathcal{Q}.\text{dequeue}()$
      \ENDWHILE

    \ENDFOR
    \STATE \textbf{Output}: Dynamically sampled frame subset $\mathcal{Q}_s$
  \end{algorithmic}
\end{algorithm}
\subsubsection{Multi-frame feature fusion}
After selecting \(N\) historical frames, the BEV features \(B_1, B_2, \dots, B_N\) are extracted from each frame using a BEV feature extraction module, while the BEV feature \(B_t\) is extracted from the current frame. To ensure spatial consistency, historical BEV features are transformed into the current coordinate frame using transformation matrices \(T_i\), resulting in transformed features \(\hat{B}_1, \hat{B}_2, \dots, \hat{B}_N\), where \(\hat{B}_i = T_i B_i\). These transformed historical features and the current BEV feature \(B_t\) are processed through an encoder network to reduce feature dimensions. The encoder output for each transformed feature is flattened to produce encoded features \(\tilde{B}_1, \tilde{B}_2, \dots, \tilde{B}_N\) for the historical frames and \(\tilde{B}_t\) for the current frame. The encoded features are processed by a GRU network as follows:
\begin{equation}
\begin{split}
h_i = \text{GRU}(\tilde{B}_i, h_{i-1}), &\quad \mathbf{p}_i = \text{MLP}(h_i) \\
B_f = \text{GRU}(\tilde{B}_t, h_N), &\quad \mathbf{p}_t = \text{MLP}(B_f)
\end{split}
\end{equation}
here, \(h_i\) is the hidden state produced by the GRU for each historical frame \(i\), based on the encoded feature \(\tilde{B}_i\) and the previous hidden state \(h_{i-1}\). The hidden state \(h_i\) is then passed through a MLP to predict the pose \(\mathbf{p}_i\) of the preceding vehicle at that specific historical frame. After processing all historical frames \(N\), the final hidden state \(h_N\) is combined with the encoded current BEV feature \(\tilde{B}_t\) to produce the fused BEV feature \(B_f\). The current preceding vehicle pose \(\mathbf{p}_t\) is then obtained by passing the fused feature \(B_f\) through an MLP. Meanwhile, the preceding vehicle's velocity \(\mathbf{v}_t\) is inferred by fusing features from historical frames selected at a fixed time interval, allowing for accurate velocity prediction based on temporal information.

\subsubsection{Planning head and loss function}
The fused feature \(B_f\) is passed through an MLP to reduce dimensionality. The compressed features are then input into a GRU-based auto-regressive trajectory point prediction network, initializing a 128-dimensional hidden state. The network takes the current vehicle position as input and, using a single-layer GRU and a linear layer, predicts differential trajectory points for the next 10 time steps \(\{ \delta w_t \}_{t=1}^{10}\). Trajectory points are computed using the recursive formula \( w_t = w_{t-1} + \delta w_t \), resulting in 10 future trajectory points spaced 0.3 seconds apart. The initial input is (0, 0), representing the vehicle-centered position in the BEV space. This method follows the waypoint prediction strategy from the TransFuser \cite{prakash2021multimodalfusiontransformerendtoend}. In addition to the three loss components mentioned earlier, we also incorporate a depth cross-entropy loss \(\mathcal{L}_d\) for depth distribution and a binary cross-entropy loss for the semantic mask \(\mathcal{L}_s\). The complete loss function is formulated as:
\begin{equation}
  \mathcal{L}(\theta) = \lambda_w \mathcal{L}_w(\theta) + \lambda_p \mathcal{L}_p(\theta) + \lambda_v \mathcal{L}_v(\theta) + \lambda_d \mathcal{L}_d(\theta) + \lambda_s \mathcal{L}_s(\theta)
\end{equation}
where \(\lambda_w\), \(\lambda_p\), \(\lambda_v\), \(\lambda_d\), and \(\lambda_s\) are the weighting factors for each loss component.

\section{Experiment}
\subsection{Real-World experimental platform and expert system}

\subsubsection{Real-World experimental platform}

\begin{figure}[t]
  \centering
  \includegraphics[width=0.45\textwidth]{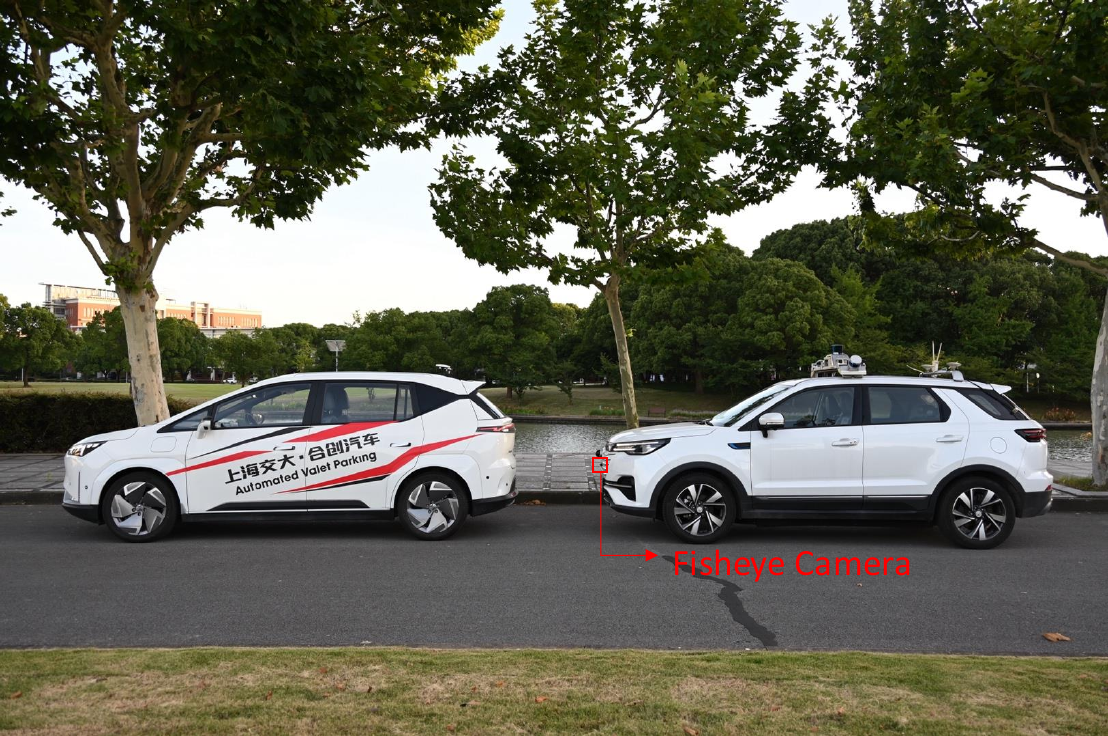}
  \caption{Real-world experimental platform.}
  \label{fig:platform}
  \vspace{-15pt}
\end{figure}

We establishes a vehicle following experimental platform as shown in Figure \ref{fig:platform}. The ego vehicle is a modified Changan CS55 E-Rock, equipped with a front-facing fisheye camera. The fisheye camera model is Senyun IM390, with a horizontal field of view of 216°, a resolution of 1920×1080, and is a mass-produced automotive sensor. We chose the fisheye camera specifically because its large horizontal FOV ensures that the preceding vehicle remains within the ego vehicle’s field of view even during high-curvature turns. This capability is crucial for maintaining robust and continuous vehicle following in scenarios with sharp turns, where traditional cameras with narrower FOVs might lose sight of the preceding vehicle. It is also equipped with a consumer-grade IMU measurement unit. The computing platform for the ego vehicle is the TW-609 vehicle computing platform based on the NVIDIA Jetson AGX Xavier. The preceding vehicle in the vehicle following system is a modified Hechuang Z03, equipped with the same IMU measurement unit as the ego vehicle. 

\subsubsection{Expert system}
In the vehicle-following experiment, the expert system relies on precise data collection and communication, as high-quality data is essential for reliable training and evaluation. In addition to the sensors involved in the above system scheme, to create a dataset for training and evaluating the vehicle following experiment, both the ego vehicle and the preceding vehicle are equipped with high-precision RTK-GPS devices to provide high-precision global localization results for the vehicles. The preceding vehicle uses an industrial computer to connect to sensors and collect data and has an outdoor wireless router installed to establish a wireless local area network for vehicle-to-vehicle (V2V) communication. Additionally, the Ego vehicle is equipped with a Pandar40 mechanical LiDAR from Hesai Technology for scenario visualization and depth ground truth generation. The LiDAR and RTK-GPS data are only used for the quantitative evaluation of the vehicle following algorithm and are not inputs to the proposed algorithm.

Our dataset was collected through the expert system, comprising a total of 30,000 frames, with 26,000 frames used for training and 4,000 frames for testing. In each frame, fisheye images serve as the input to the algorithm, while depth maps of the fisheye images, semantic masks, vehicle poses, velocity of the preceding vehicle, and planned trajectories of the ego vehicle are used as the supervised ground truth for the algorithm's outputs. Our expert system is a well-designed vehicle following system. In the expert system's vehicle following algorithm, the fisheye camera depth ground truth is obtained by projecting the LiDAR point cloud into the fisheye camera coordinate system. The perception ground truth of the preceding vehicle is calculated based on the relative poses obtained from high-precision RTK GPS units installed on both vehicles. The planned trajectories for the next 3 seconds for the ego vehicle consist of 10 trajectory points with a time interval of 0.3 seconds.

\subsection{Closed-loop experiment}
\subsubsection{Realistic route experiment}

\begin{figure}[t]
  \centering
  \includegraphics[width=0.45\textwidth]{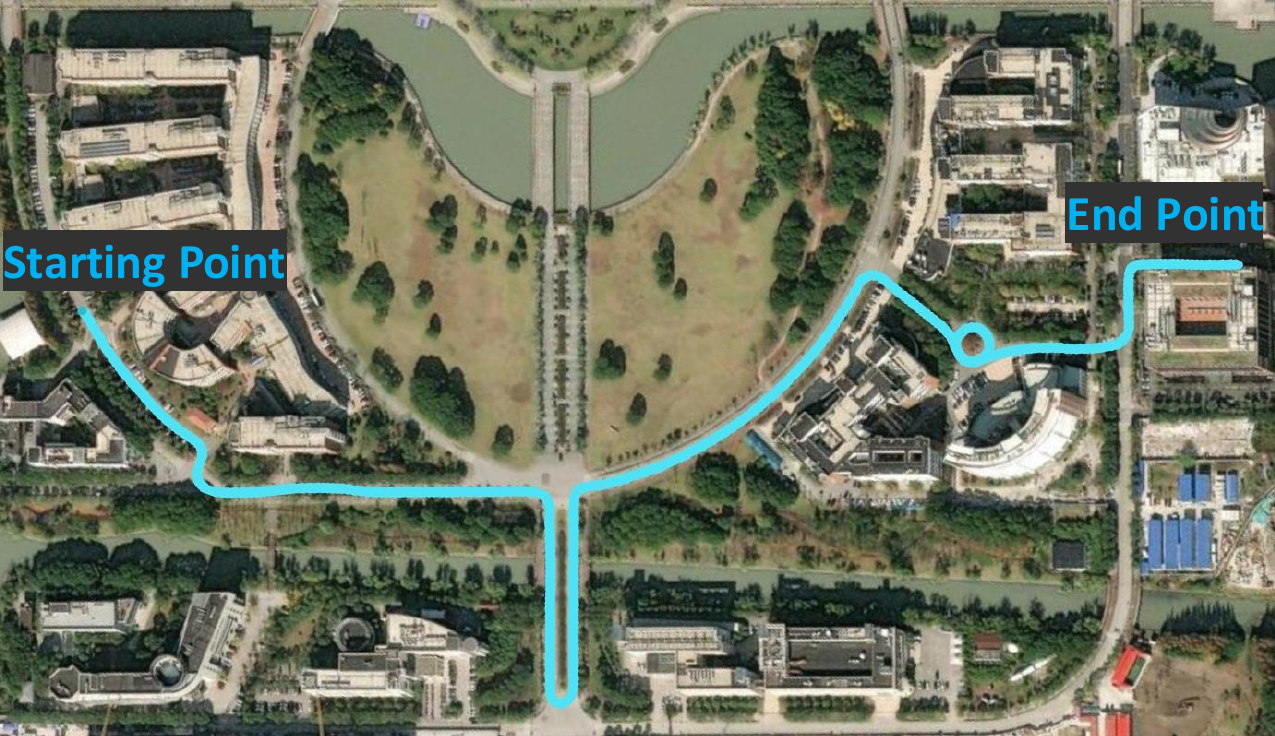}
  \caption{Realistic route for closed-loop experiment: A 1.5 km route containing diverse traffic scenarios (straight roads, curves, intersections, and roundabouts), where the preceding vehicle maintained a velocity below 6 m/s, an acceleration within 1.5 m/s², a fixed following distance of 4 meters, and a time gap of 0.5 seconds.}
  \label{fig:test-route}
\end{figure}

\begin{figure*}[ht]
  \centering
  \subfloat[Rule-based method]{
    \includegraphics[width=0.32\textwidth]{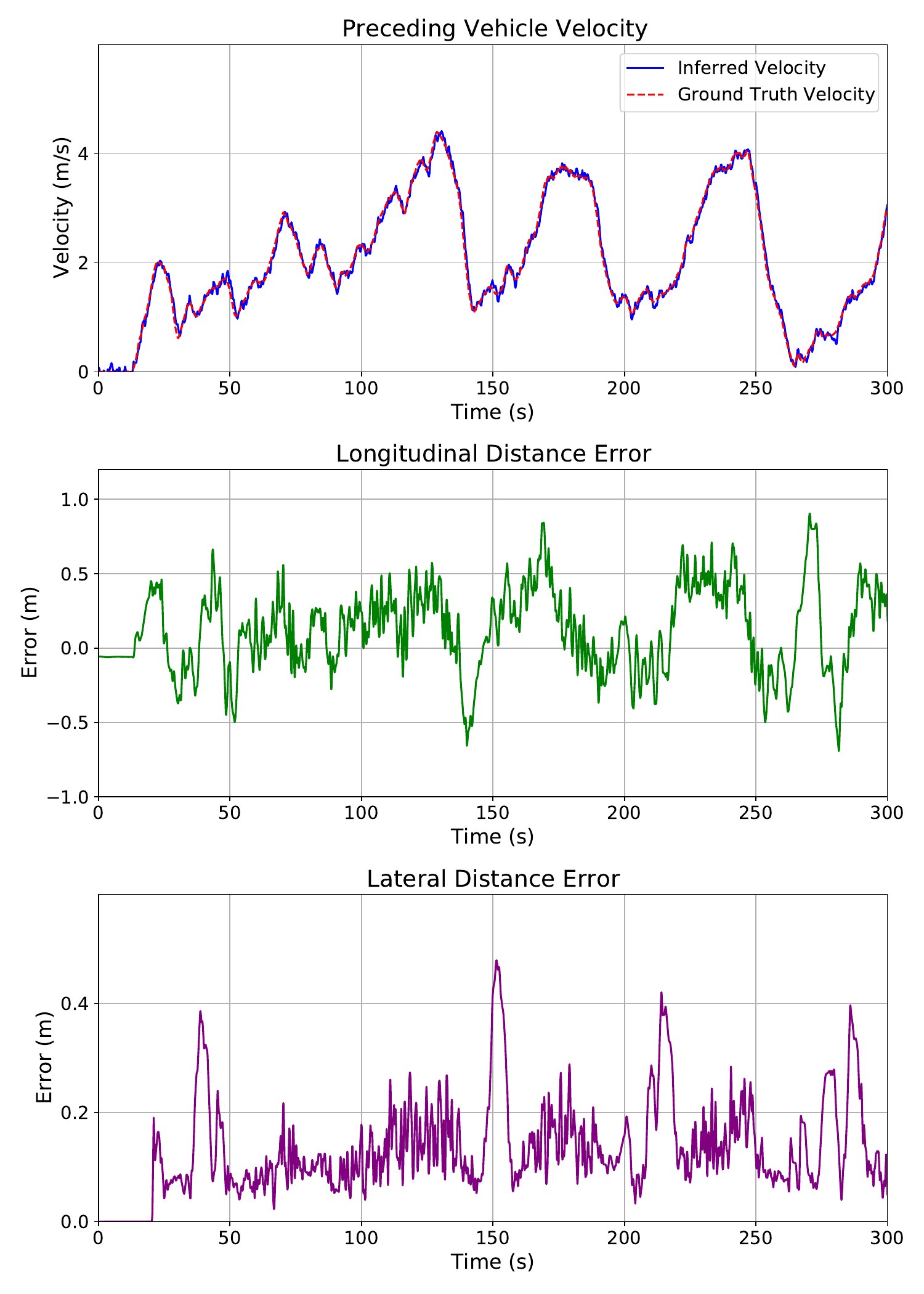}
    \label{fig:closed-loop-a}
  }
  \subfloat[Ours (E2E)]{
    \includegraphics[width=0.32\textwidth]{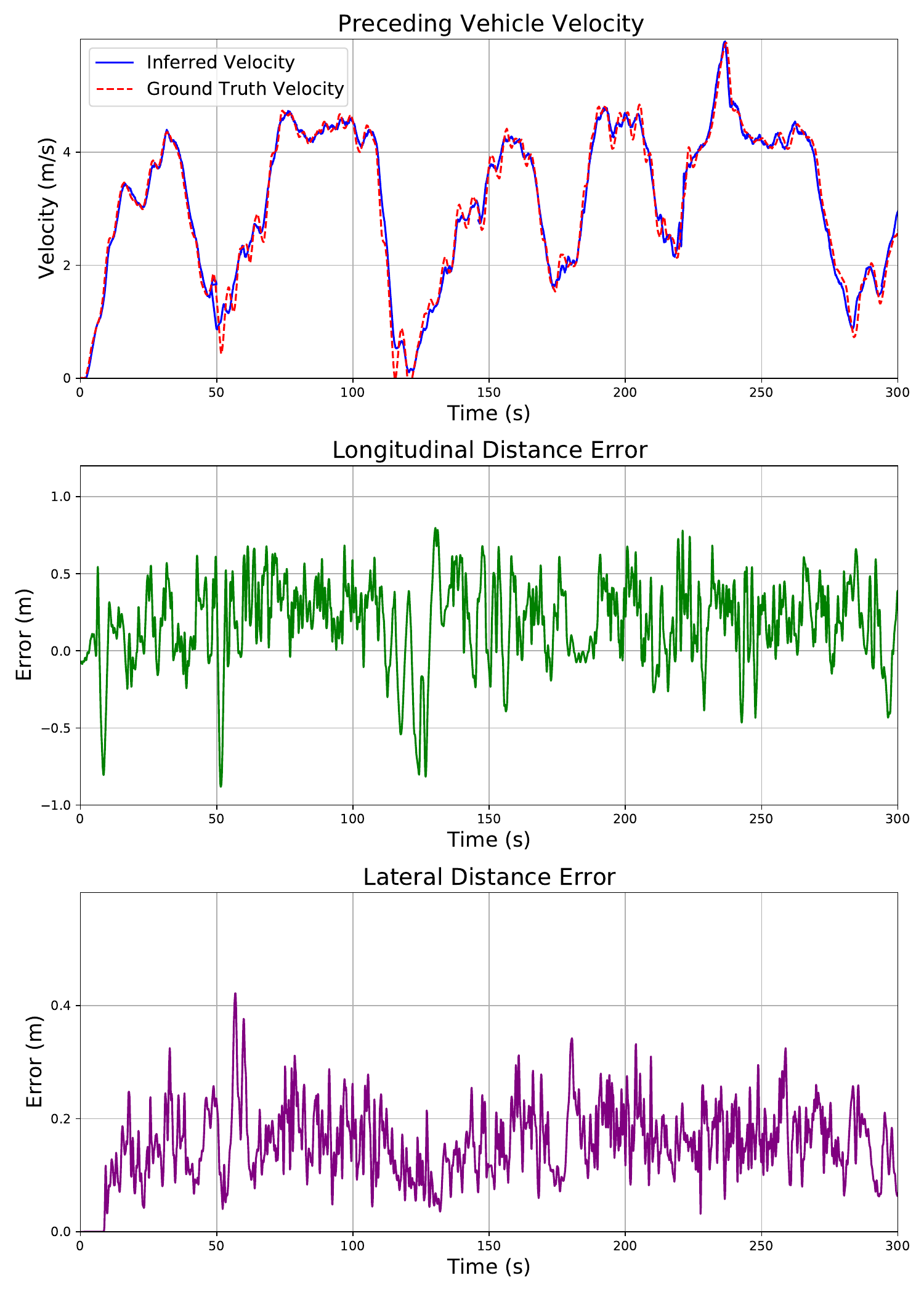}
    \label{fig:closed-loop-b}
  }
  \subfloat[Multi-stage method]{
    \includegraphics[width=0.32\textwidth]{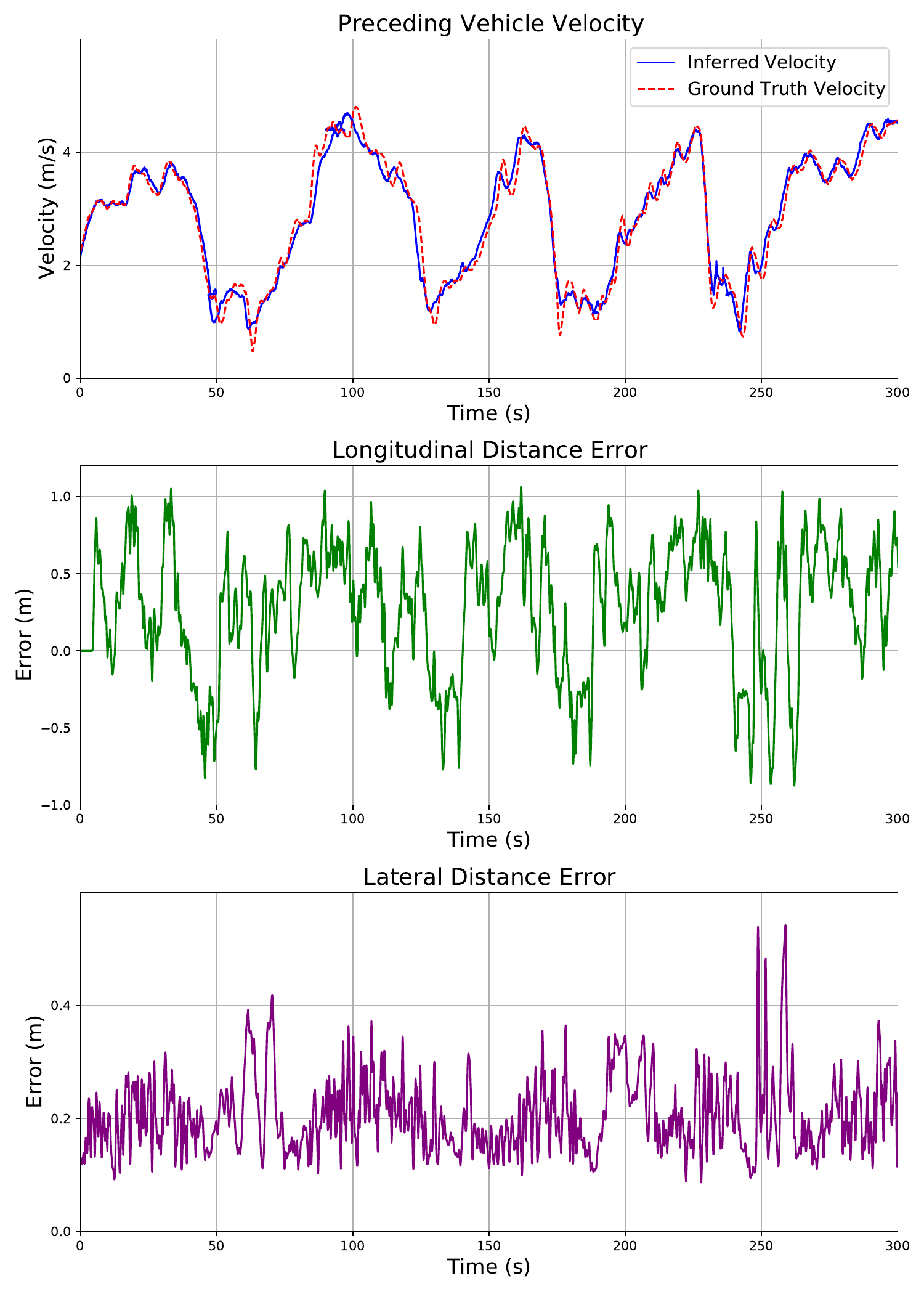}
    \label{fig:closed-loop-c}
  }
  \caption{Closed-loop experiment results for different methods: The Rule-based method detects the position of the preceding vehicle using a fisheye camera and millimeter-wave radar, while also leveraging V2V devices to directly obtain the preceding vehicle's wheel speed, resulting in more accurate velocity estimation. In contrast, this work relies solely on a single fisheye camera.}
  \label{fig:closed-loop}
\end{figure*}

We conducted a closed-loop experiment on a diverse route to evaluate the performance of our algorithm in real-world scenarios, as illustrated in Figure \ref{fig:test-route}. The experiment was conducted under different weather conditions, to assess the algorithm's robustness. Figure \ref{fig:closed-loop} shows the position and velocity of the preceding vehicle, as well as the longitudinal and lateral following errors throughout the test.

In the experiment, we compared our proposed end-to-end method Ours (E2E) with two other approaches. The first approach, the Rule-based method that uses a fisheye camera and millimeter-wave radar for perception to detect the position of the preceding vehicle and obtains the wheel speed of the preceding vehicle through V2V devices. Then it applies rule-based planning and MPC controllers to generate the trajectory of the ego vehicle. The second approach is Multi-stage method, where the perception network is trained first independently, followed by training the planning network based on the output of the perception network. The perception results are explicitly passed between stages, making it a Multi-stage process.

The comparison results presented in Table \ref{tab:comparison_algorithms} show that Ours (E2E) achieved an average longitudinal error of 0.22 meters and an average lateral error of 0.14 meters, outperforming both the Rule-based method and Multi-stage approaches. These results further validate the effectiveness of our method, where joint optimization of perception and planning leads to better performance in real-world scenarios.

\subsubsection{Lateral accuracy experiment}
\begin{figure*}[t]
  \centering
  \subfloat[{{\normalsize Roundabout}}]{\includegraphics[height=5cm]{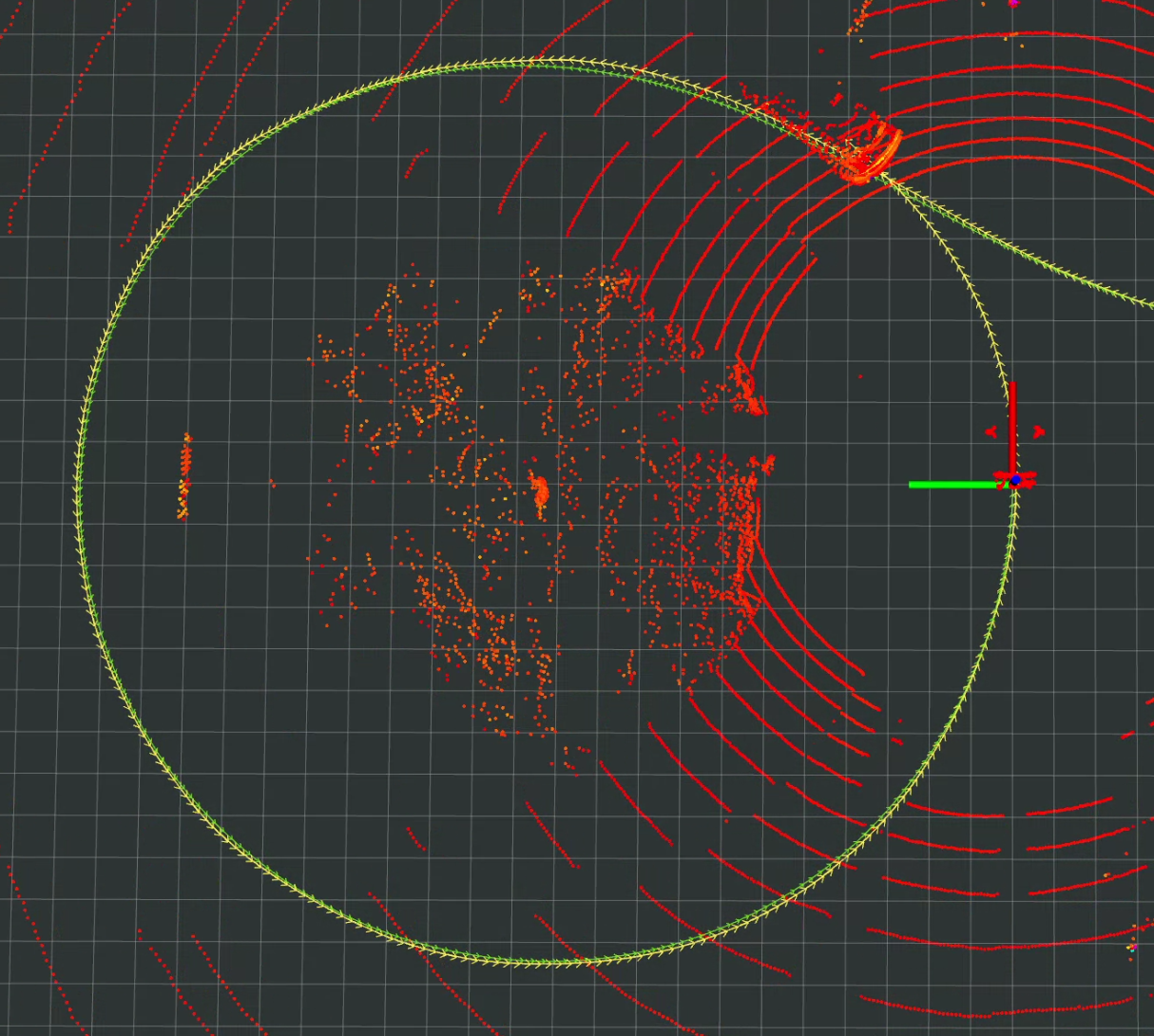}} \hspace{5mm}
  \subfloat[{\normalsize Right-angle curve}]{\includegraphics[height=5cm]{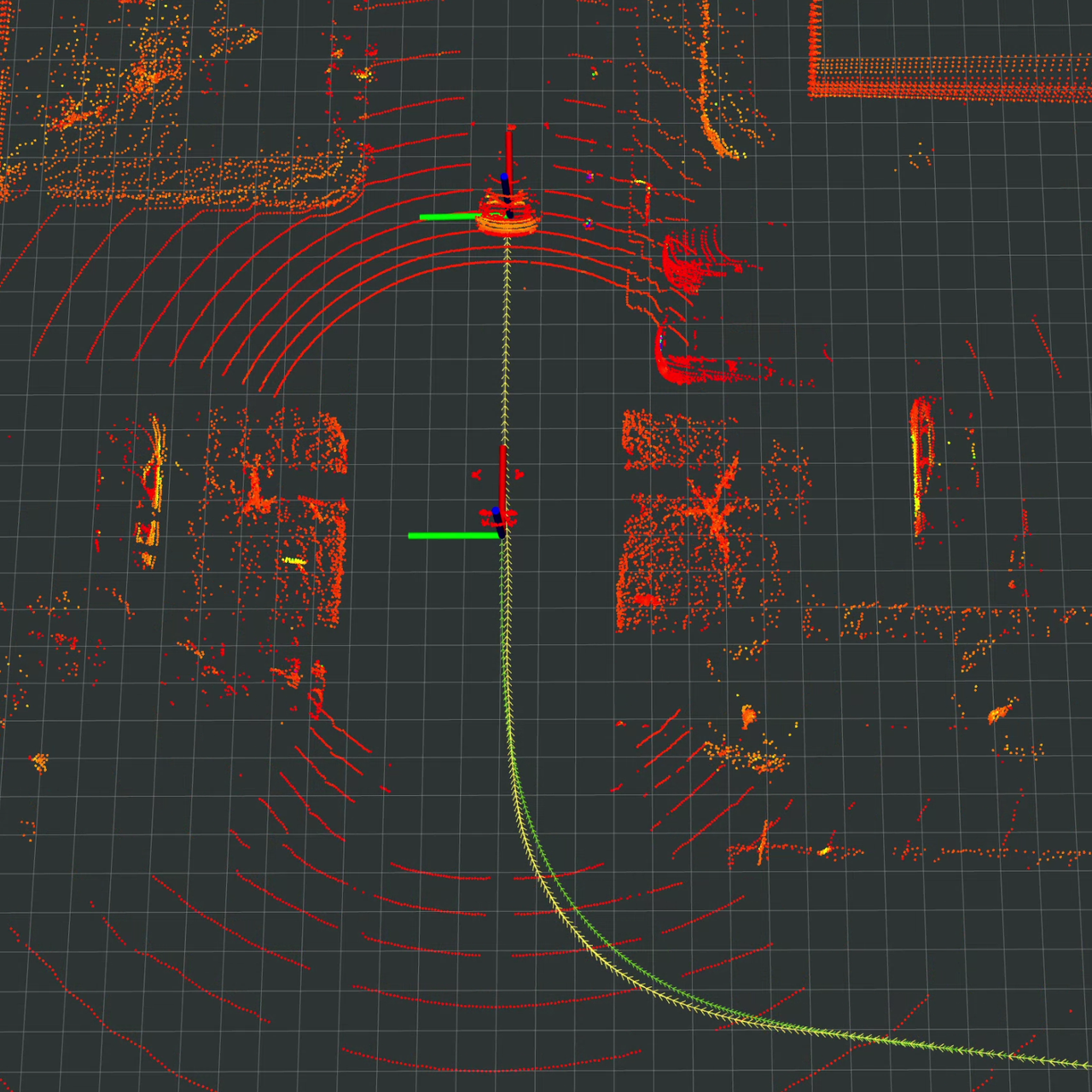}} \hspace{5mm}
  \subfloat[{\normalsize U-turn}]{\includegraphics[height=5cm]{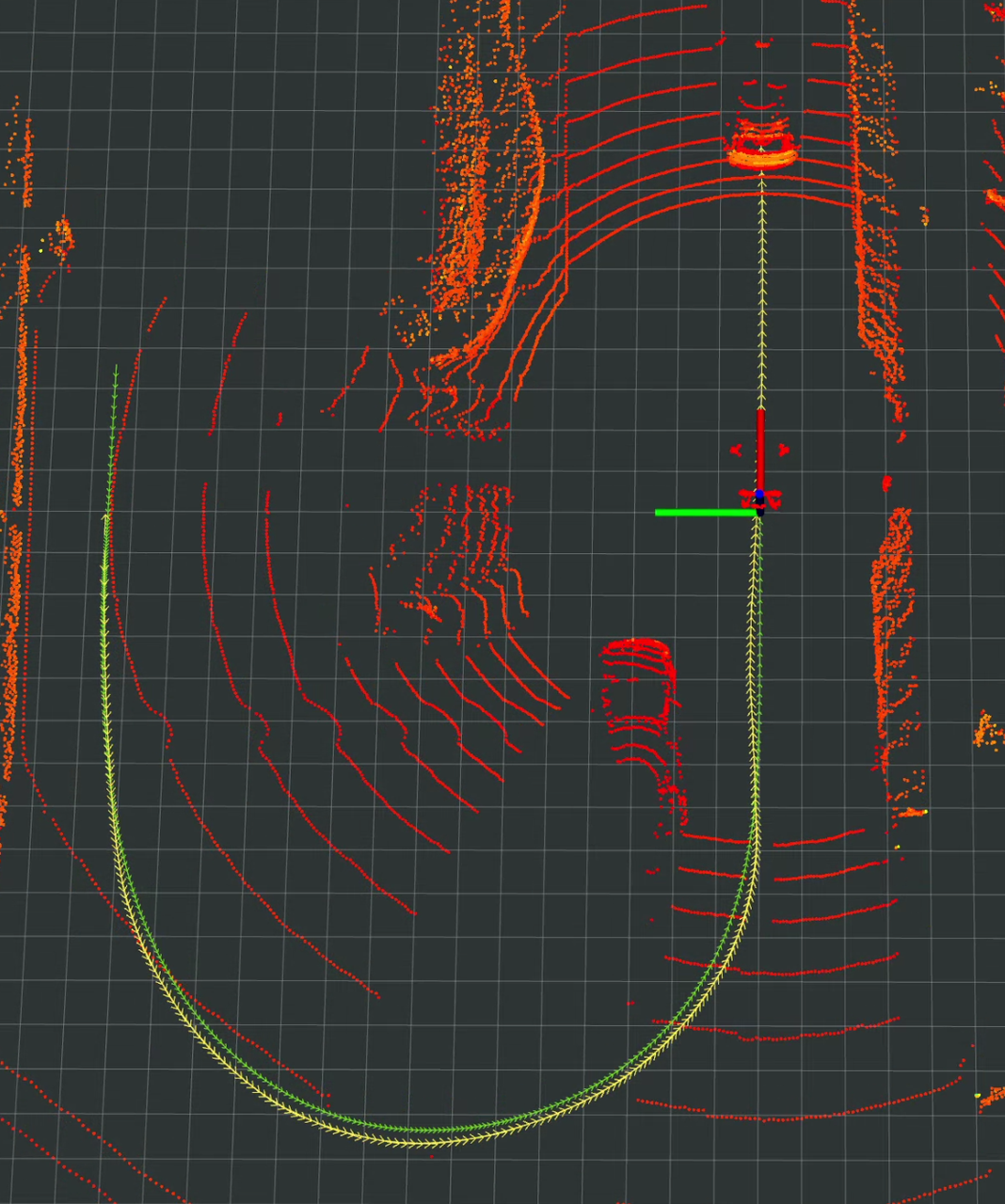}}
  \caption{Qualitative results for Ours (E2E) method in different driving scenarios: (a) Roundabout, (b) Right-angle curve, (c) U-turn. The yellow curve represents the trajectory of the preceding vehicle, and the green curve represents the trajectory of the ego vehicle. The side length of each white grid corresponds to 1 meter. It can be observed from the figures that the trajectory of the ego vehicle closely overlaps with that of the preceding vehicle.}
  \label{fig:qualitative-lateral}
\end{figure*}

We conducted lateral tracking accuracy tests in roundabout, right-angle curve, and U-turn scenarios. The quantitative results, as shown in Table \ref{tab:lateral_accuracy}, are compared with the Multi-stage method and demonstrate that our algorithm consistently outperforms the Multi-stage method in terms of lateral tracking accuracy. The corresponding qualitative results for Ours (E2E) method in these scenarios can be seen in Figure \ref{fig:qualitative-lateral}.

\begin{table}[ht]
\centering
\caption{Comparison of Algorithms for Following Errors}
\label{tab:comparison_algorithms}
\small 
\setlength{\tabcolsep}{3pt}
\begin{tabular}{lcccc}
\toprule
\textbf{Algorithm} & \textbf{Avg. Long.$\downarrow$} & \textbf{Max Long.$\downarrow$} & \textbf{Avg. Lat.$\downarrow$} & \textbf{Max Lat.$\downarrow$} \\
\midrule
Rule-based         & 0.24               & \textbf{0.89}      & 0.16              & 0.54              \\
Multi-stage        & 0.41               & 1.09               & 0.25              & 0.60              \\ 
Ours (E2E)         & \textbf{0.22}      & 0.94               & \textbf{0.14}     & \textbf{0.42}     \\
\bottomrule
\end{tabular}
\\\vspace{1mm}
{\small Avg. = Average, Max = Maximum (errors); Long.=Longitudinal, Lat.=Lateral (in meters). The Rule-based method uses a fisheye camera and millimeter-wave radar to detect the preceding vehicle’s position and V2V devices for wheel speed, whereas Ours utilizes only a single fisheye camera.}
\end{table}

\begin{table}[ht]
\centering
\caption{Lateral Tracking Accuracy Comparison}
\label{tab:lateral_accuracy}
\small 
\setlength{\tabcolsep}{3pt}
\begin{tabular}{lcccc}
\toprule
\textbf{Scenario} & \multicolumn{2}{c}{\textbf{Ours}} & \multicolumn{2}{c}{\textbf{Multi-stage}} \\ 
                  & \textbf{Avg. (m)} & \textbf{Max. (m)} & \textbf{Avg. (m)} & \textbf{Max. (m)} \\ \midrule
Roundabout        & \textbf{0.11}     & \textbf{0.25}     & 0.20              & 0.49              \\ 
Right-angle curve & \textbf{0.11}     & \textbf{0.39}     & 0.36              & 0.53              \\ 
U-turn            & \textbf{0.13}     & \textbf{0.29}     & 0.26              & 0.58              \\ 
\bottomrule
\end{tabular}
\end{table}

\begin{table}[h]
\centering
\caption{Performance Comparison of Methods}
\label{tab:performance_comparison}
\small 
\setlength{\tabcolsep}{3pt}
\begin{tabular}{lcccccc}
\toprule
\textbf{Method} & \textbf{RT (ms)$\downarrow$} & \textbf{UJ (/km)$\downarrow$} & \multicolumn{2}{c}{\textbf{Max Err. (m)}$\downarrow$} & \multicolumn{2}{c}{\textbf{PBD (m)}$\uparrow$} \\
\cmidrule(lr){4-5} \cmidrule(lr){6-7}
 &  &  & \textbf{3.5} & \textbf{5} & \textbf{3.5} & \textbf{5} \\\midrule
Multi-stage    & 124 & 15 & 1.96 & 2.78 & 2.28 & 2.08 \\
Ours (E2E)     & \textbf{87} & \textbf{3} & \textbf{1.93} & \textbf{2.00} & \textbf{2.5} & \textbf{3.0} \\
\bottomrule
\end{tabular}
\\\vspace{1mm}
{\small RT = Response Time, UJ = Uncomfortable Jerk, Max Err. = Maximum Error, PBD = Post-braking Distance. 3.5 and 5 m/s are pre-braking speeds of the preceding vehicle.}
\end{table}

\begin{figure}[t]
  \centering
  \subfloat[10 km/h to 20 km/h]{\includegraphics[width=0.25\textwidth]{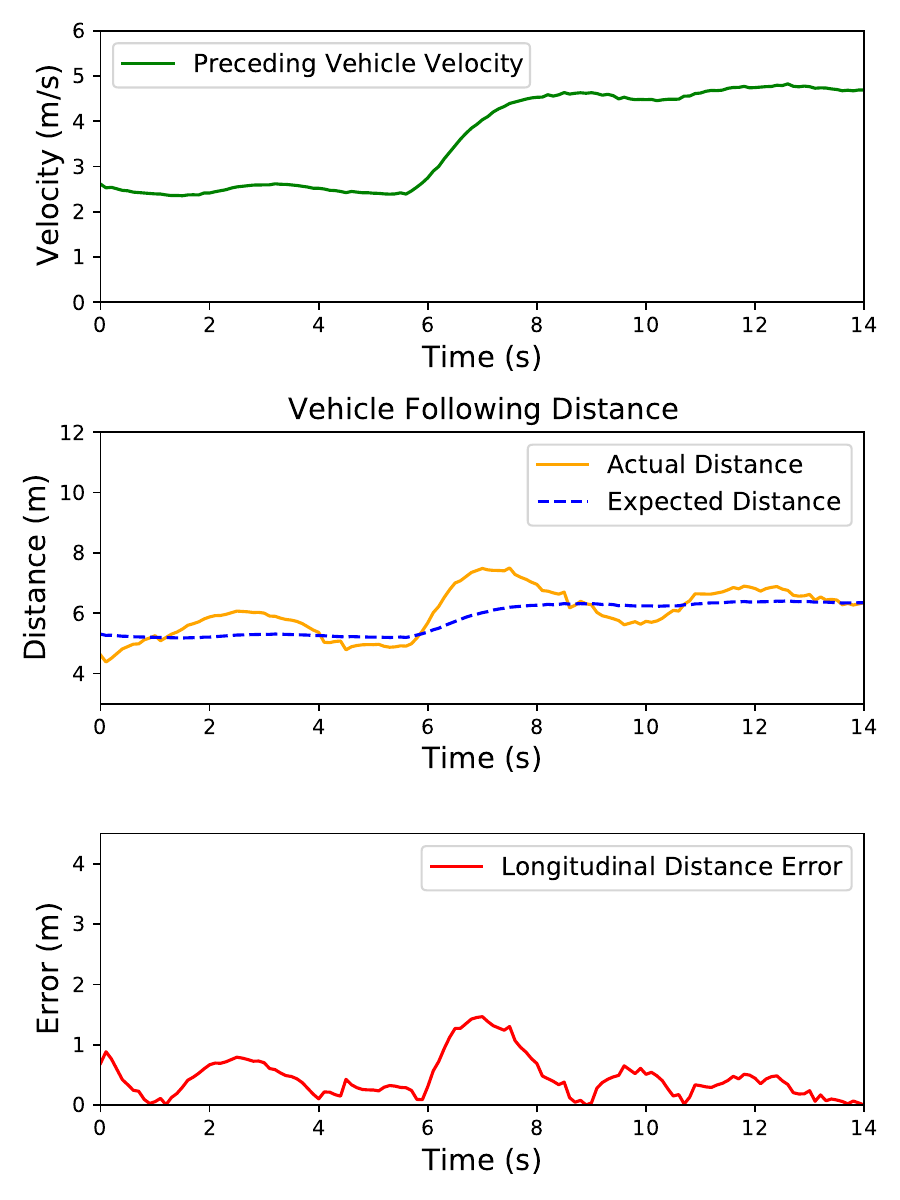}}
  \subfloat[0 km/h to 20 km/h]{\includegraphics[width=0.25\textwidth]{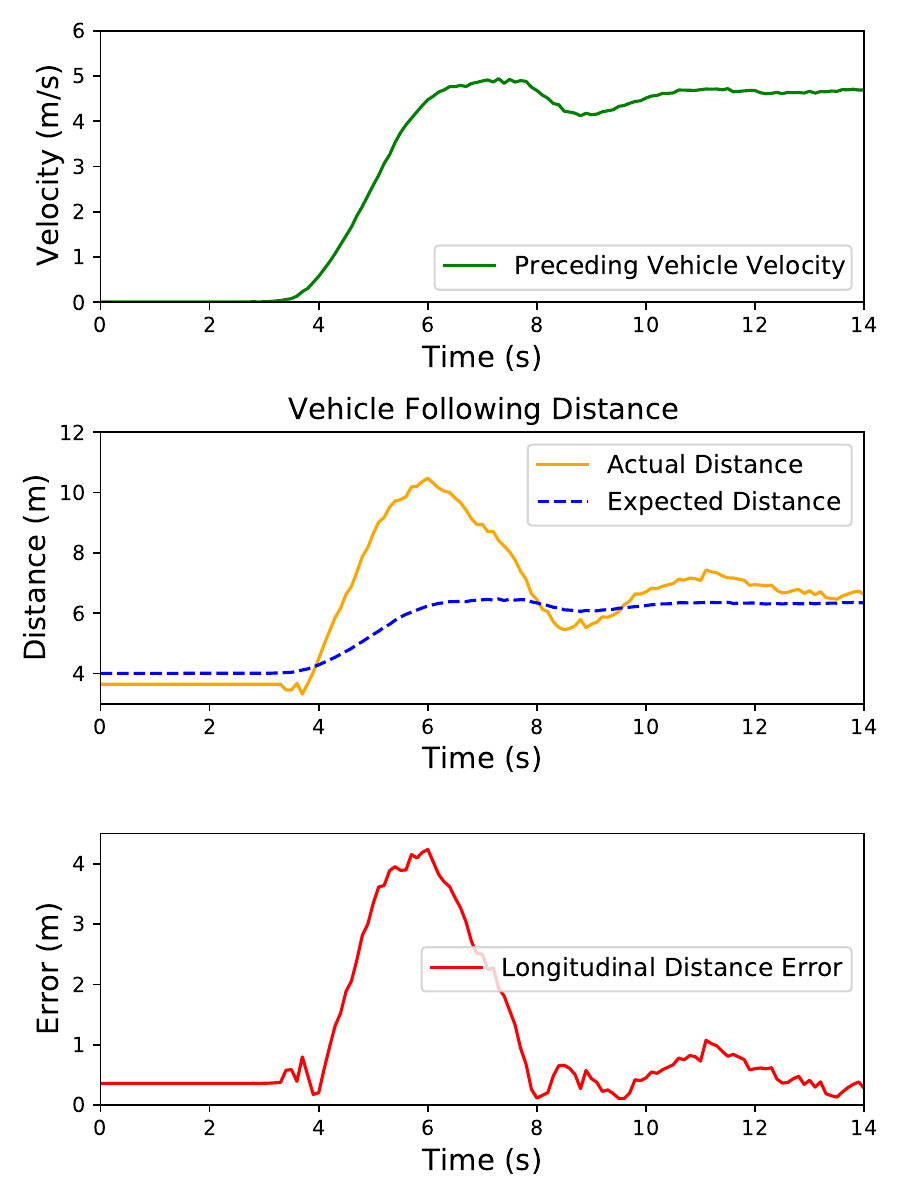}}
  \caption{Sudden acceleration scenarios and tracking performance: (a) 10 km/h to 20 km/h, where the maximum longitudinal error of our method was 1.46 m, demonstrating fast and stable reactions; (b) 0 km/h to 20 km/h, where our method achieved a maximum longitudinal error of 4.23 m and successfully maintained tracking, while the Multi-stage method failed.}
  \label{fig:acc}
  \vspace{-15pt}
\end{figure}

\subsubsection{Longitudinal accuracy experiment}
We analyze the longitudinal error in scenarios where the preceding vehicle underwent rapid deceleration, specifically focusing on emergency braking. The quantitative results for the emergency braking scenarios are summarized in Table \ref{tab:performance_comparison}, which shows that Ours (E2E) method outperforms the Multi-stage method. Based on extensive experimental results, we also evaluated the metric of response time, defined as the duration the system takes to execute an output action after receiving an input signal. These results illustrate that our (End-to-end) method achieves faster response times and improved accuracy compared to the Multi-stage method.

Furthermore, we evaluated the frequency of uncomfortable jerks per kilometer, which highlights the effectiveness of our method in mitigating the impact of perception errors. In contrast, the Multi-stage method, being more sensitive to perception inaccuracies, often propagates these errors to the trajectory planning and control stages, leading to frequent abrupt maneuvers that compromise passenger comfort.

\begin{figure*}[ht]
  \centering
  \subfloat[Narrow Driving Space]{
    \includegraphics[width=0.23\textwidth]{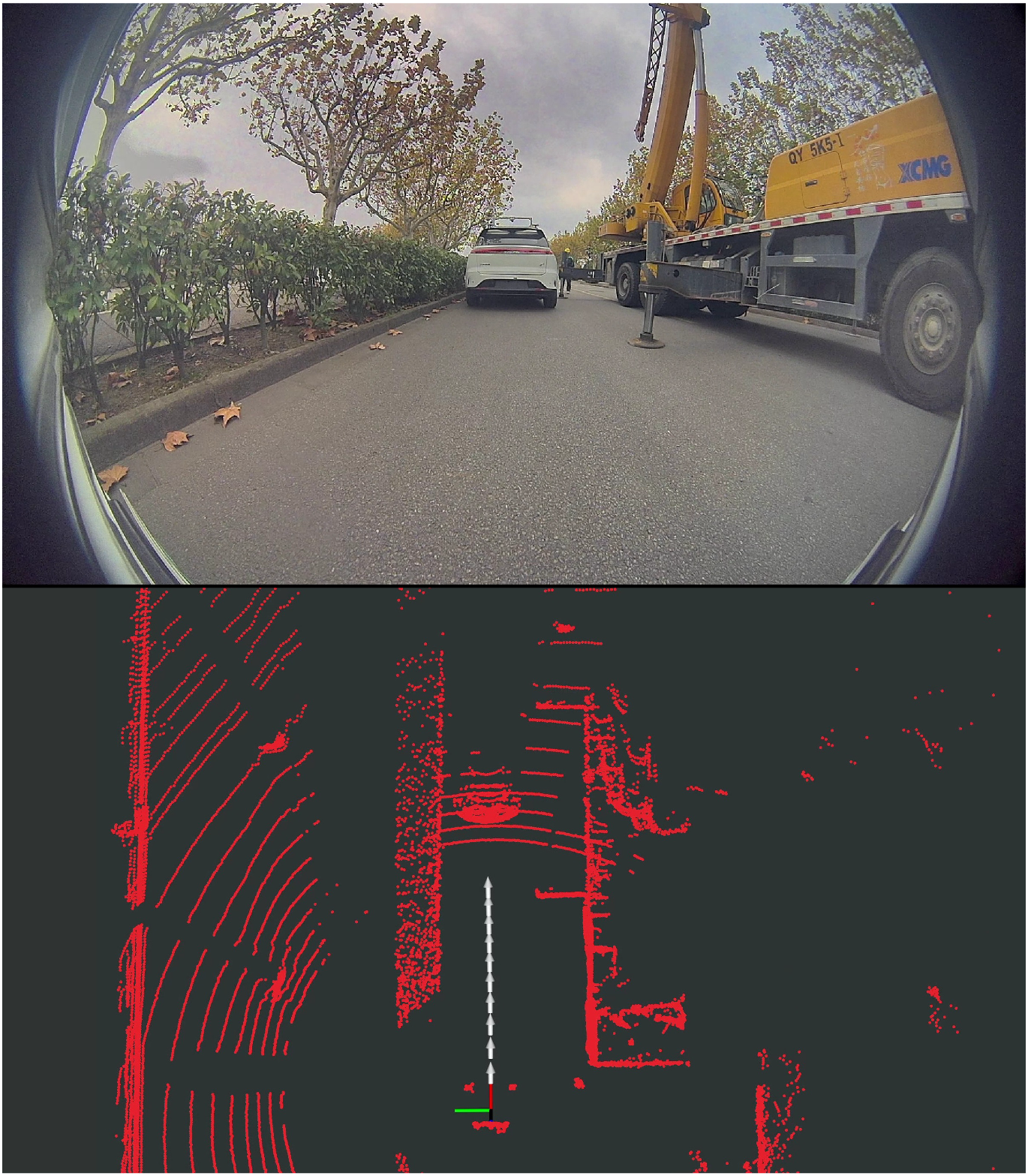}
    \label{fig:narrow}
  }
  \subfloat[Low-light Environment]{
    \includegraphics[width=0.23\textwidth]{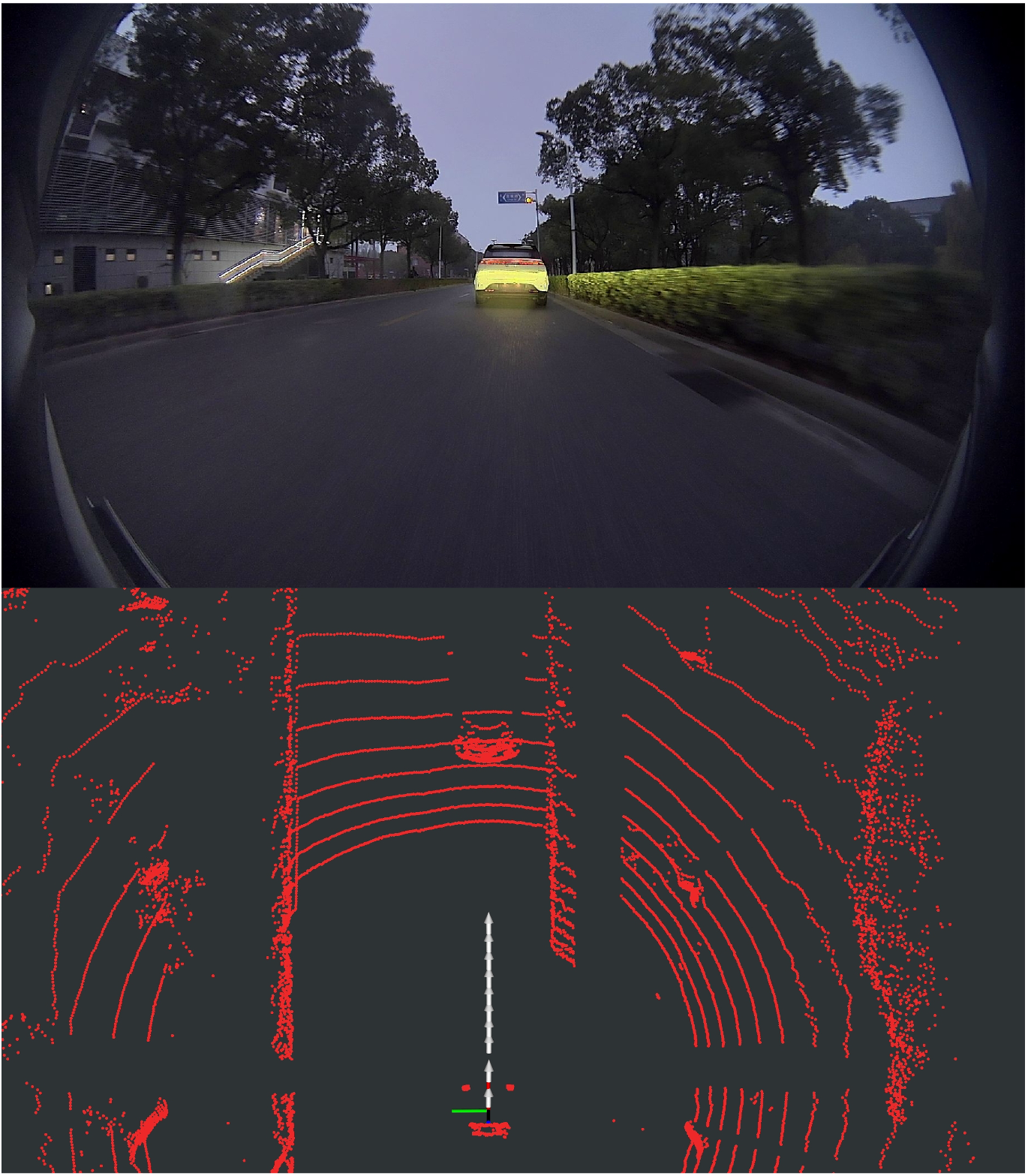}
    \label{fig:night}
  }
  \subfloat[Dense Vehicles]{
    \includegraphics[width=0.23\textwidth]{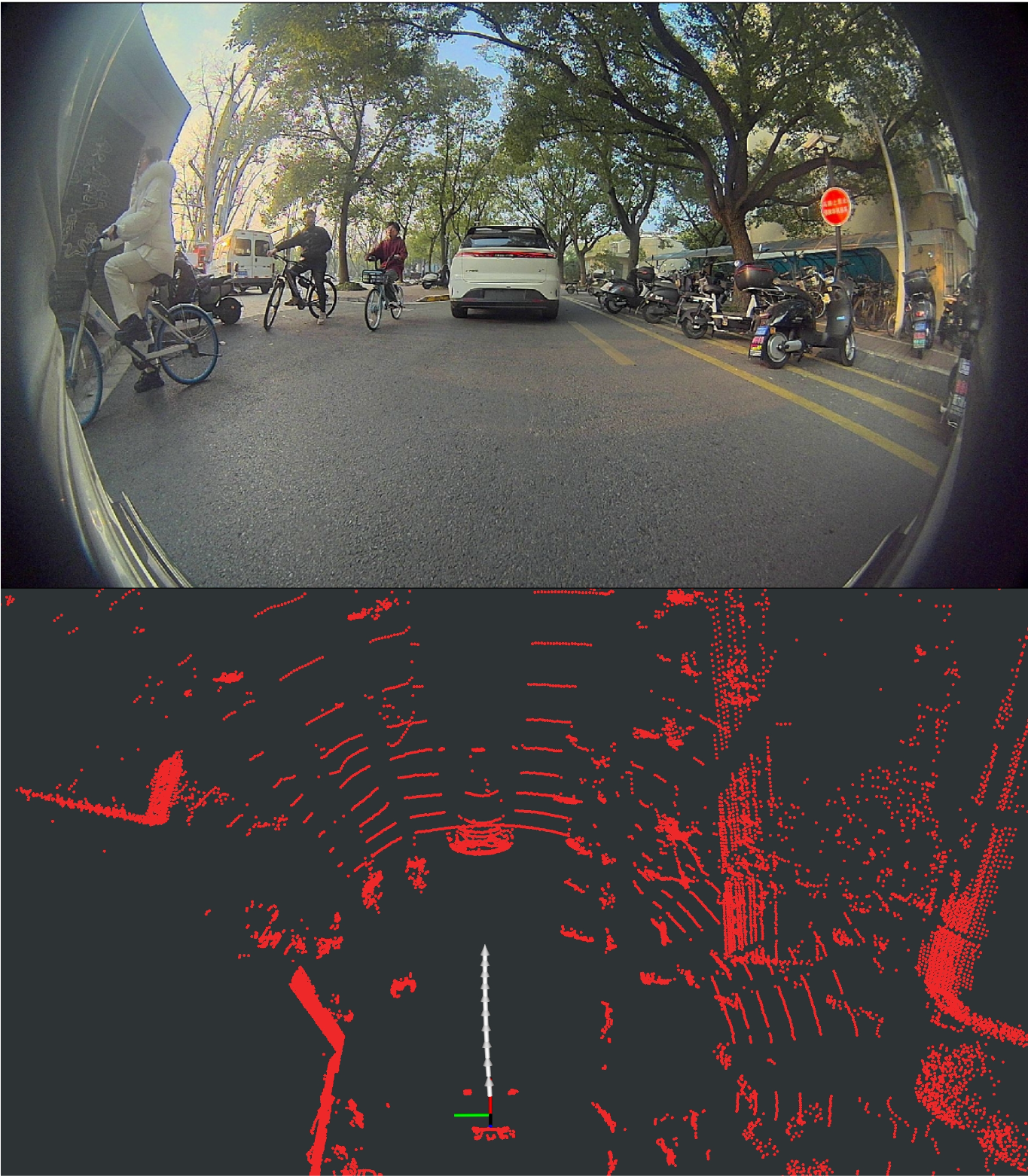}
    \label{fig:people}
  }
  \subfloat[Rainy Condition]{
    \includegraphics[width=0.23\textwidth]{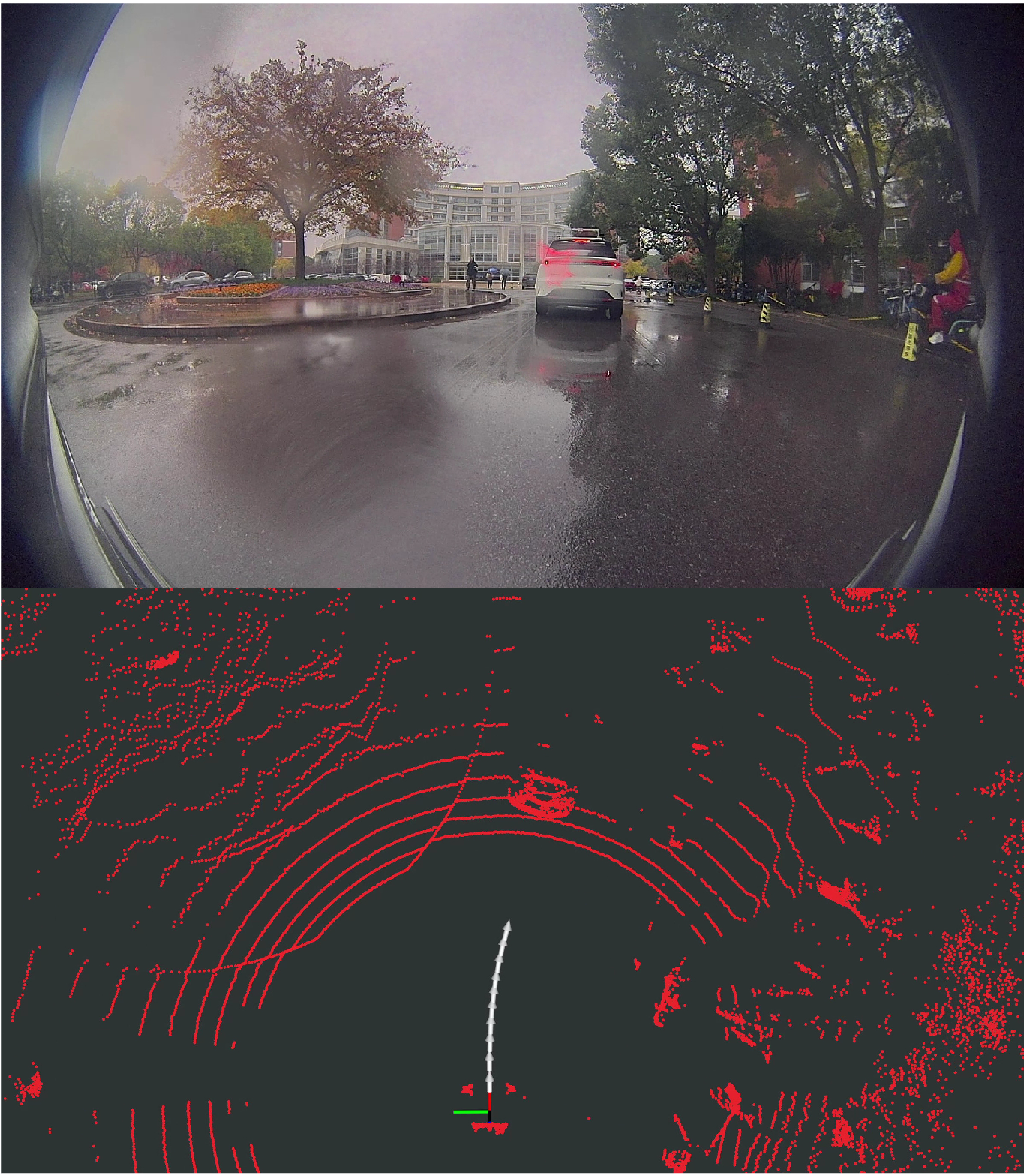}
    \label{fig:rain}
  }
  \caption{Existing multi-stage methods often fail in the following challenging real-world scenarios: (a) narrow driving space with only 20 cm clearance from obstacles on one side, (b) low-light environments, (c) dense non-motorized vehicle environments, and (d) rainy conditions where raindrops on the fisheye lens lead to image distortions. These scenarios present significant challenges to autonomous driving systems, while our approach demonstrates robust performance. The complete experimental process can be seen in the accompanying video.}
  \label{fig:robust-scenarios}
\end{figure*}

\subsubsection{Challenging Scenarios Experiment}
In this section, we evaluate the performance of our method in challenging scenarios, demonstrating its enhanced robustness compared to the Multi-stage method. For example, in the sudden acceleration scenario (Figure \ref{fig:acc}) where the preceding vehicle was initially stationary and then accelerated to 20 km/h, our method achieved a maximum longitudinal error of 4.23 m and successfully maintained tracking, whereas the Multi-stage method failed to track. In addition to these quantitative results, we tested our method in challenging real-world scenarios, including scenarios with dense non-motorized vehicle traffic, narrow driving spaces, rainy conditions, and low-light environments. Moreover, as illustrated in Figure \ref{fig:robust-scenarios}, our method demonstrated reliable decision making and path planning in these challenging scenarios, successfully navigating complex environments where the Multi-stage method struggled. The results demonstrate the reliability of our end-to-end approach in maintaining vehicle tracking under complex conditions.

\section{Discussion}
This research presents an innovative method for vehicle following, utilizing only a monocular fisheye camera for perception and trajectory planning in an end-to-end framework. The system successfully addresses the challenges of autonomous vehicle following in diverse and unstructured road environments, providing a cost-effective alternative to traditional methods that rely on expensive multi-sensor setups such as LiDAR or GPS.

By incorporating a dynamic sampling mechanism and semantic mask, the end-to-end method was successfully deployed in real vehicles with closed-loop validation, demonstrating excellent performance in both longitudinal and lateral control, particularly in scenarios involving frequent stops, starts, and high-curvature turns. Compared to traditional multi-stage methods, it offers higher accuracy and robustness, making it a scalable solution for autonomous vehicle platooning. 

Future work will focus on ensuring the stability of the system as the number of vehicles in the platoon increases, while continuing to optimize real-time performance.

\bibliography{ref}

@article{dey2015review,
  title     = {A review of communication, driver characteristics, and controls aspects of cooperative adaptive cruise control (CACC)},
  author    = {Dey, K. C. and Yan, L. and Wang, X. and Wang, Y. and Shen, H. and Chowdhury, M. et al.},
  journal   = {IEEE Transactions on Intelligent Transportation Systems},
  volume    = {17},
  number    = {2},
  pages     = {491--509},
  year      = {2015},
  publisher = {IEEE}
}

@article{9852980,
  title={A review of connected and automated vehicle platoon merging and splitting operations},
  author={Li, Qianwen and Chen, Zhiwei and Li, Xiaopeng},
  journal={IEEE Transactions on Intelligent Transportation Systems},
  volume={23},
  number={12},
  pages={22790--22806},
  year={2022},
  publisher={IEEE}
}

@article{7970188,
  title={Platoon control of connected vehicles from a networked control perspective: Literature review, component modeling, and controller synthesis},
  author={Li, Shengbo Eben and Zheng, Yang and Li, Keqiang and Wang, Le-Yi and Zhang, Hongwei},
  journal={IEEE Transactions on Vehicular Technology},
  year={2017},
  publisher={IEEE}
}

@article{1998,
  title={Vehicle following control in lateral direction for platooning},
  author={Fujioka, Takehiko and Omae, Manabu},
  journal={Vehicle System Dynamics},
  volume={29},
  number={S1},
  pages={422--437},
  year={1998},
  publisher={Taylor \& Francis}
}

@article{8598907,
  title={An integrated longitudinal and lateral vehicle following control system with radar and vehicle-to-vehicle communication},
  author={Wei, Shouyang and Zou, Yuan and Zhang, Xudong and Zhang, Tao and Li, Xiaoliang},
  journal={IEEE Transactions on Vehicular Technology},
  volume={68},
  number={2},
  pages={1116--1127},
  year={2019},
  publisher={IEEE}
}

@article{solyom2013performance,
  title={Performance limitations in vehicle platoon control},
  author={Solyom, Stefan and Coelingh, Erik},
  journal={IEEE Intelligent Transportation Systems Magazine},
  volume={5},
  number={4},
  pages={112--120},
  year={2013},
  publisher={IEEE}
}

@article{10106476,
  title={State of the art: Ongoing research in assessment methods for lane keeping assistance systems},
  author={Wei, Sijie and Pfeffer, Peter E and Edelmann, Johannes},
  journal={IEEE Transactions on Intelligent Vehicles},
  year={2023},
  publisher={IEEE}
}

@inproceedings{1570755,
  title={Nonlinear control for urban vehicles platooning, relying upon a unique kinematic GPS},
  author={Bom, Jonathan and Thuilot, Beno{\^\i}t and Marmoiton, Fran{\c{c}}ois and Martinet, Philippe},
  booktitle={Proceedings of the 2005 IEEE International Conference on Robotics and Automation (ICRA)},
  pages={4138--4143},
  year={2005},
  organization={IEEE}
}

@article{chen2024endtoendautonomousdrivingchallenges,
  title={End-to-end autonomous driving: Challenges and frontiers},
  author={Chen, Li and Wu, Penghao and Chitta, Kashyap and Jaeger, Bernhard and Geiger, Andreas and Li, Hongyang},
  journal={IEEE Transactions on Pattern Analysis and Machine Intelligence},
  year={2024},
  publisher={IEEE}
}

@article{10258330,
  title={Recent advancements in end-to-end autonomous driving using deep learning: A survey},
  author={Chib, Pranav Singh and Singh, Pravendra},
  journal={IEEE Transactions on Intelligent Vehicles},
  year={2023},
  publisher={IEEE}
}

@inproceedings{9709920,
  title={Learning to drive from a world on rails},
  author={Chen, Dian and Koltun, Vladlen and Kr{\"a}henb{\"u}hl, Philipp},
  booktitle={Proceedings of the IEEE/CVF International Conference on Computer Vision (ICCV)},
  pages={15590--15599},
  year={2021}
}

@inproceedings{peng2021safedrivingexpertguided,
  title={Safe driving via expert guided policy optimization},
  author={Peng, Zhenghao and Li, Quanyi and Liu, Chunxiao and Zhou, Bolei},
  booktitle={Conference on Robot Learning (CoRL)},
  pages={1554--1563},
  year={2022},
  organization={PMLR}
}

@inproceedings{Dosovitskiy17,
  title={CARLA: An open urban driving simulator},
  author={Dosovitskiy, Alexey and Ros, German and Codevilla, Felipe and Lopez, Antonio and Koltun, Vladlen},
  booktitle={Conference on robot learning (CoRL)},
  pages={1--16},
  year={2017},
  organization={PMLR}
}

@misc{li2022efficientlearningsafedriving,
  title         = {Efficient learning of safe driving policy via Human-AI copilot optimization},
  author        = {Quanyi Li and Zhenghao Peng and Bolei Zhou},
  year          = {2022},
  eprint        = {2202.10341},
  archiveprefix = {arXiv},
  primaryclass  = {cs.LG},
  url           = {https://arxiv.org/abs/2202.10341}
}

@inproceedings{prakash2021multimodalfusiontransformerendtoend,
  title={Multi-modal fusion transformer for end-to-end autonomous driving},
  author={Prakash, Aditya and Chitta, Kashyap and Geiger, Andreas},
  booktitle={Proceedings of the IEEE/CVF conference on computer vision and pattern recognition (CVPR)},
  pages={7077--7087},
  year={2021}
}

@inproceedings{chen2022learningvehicles,
  title={Learning from all vehicles},
  author={Chen, Dian and Kr{\"a}henb{\"u}hl, Philipp},
  booktitle={Proceedings of the IEEE/CVF Conference on Computer Vision and Pattern Recognition (CVPR)},
  pages={17222--17231},
  year={2022}
}

@article{wen2020fightingcopycatagentsbehavioral,
  title={Fighting copycat agents in behavioral cloning from observation histories},
  author={Wen, Chuan and Lin, Jierui and Darrell, Trevor and Jayaraman, Dinesh and Gao, Yang},
  journal={Advances in Neural Information Processing Systems},
  volume={33},
  pages={2564--2575},
  year={2020}
}

@misc{wen2021keyframefocusedvisualimitationlearning,
  title         = {Keyframe-focused visual imitation learning},
  author        = {Chuan Wen and Jierui Lin and Jianing Qian and Yang Gao and Dinesh Jayaraman},
  year          = {2021},
  eprint        = {2106.06452},
  archiveprefix = {arXiv},
  primaryclass  = {cs.LG},
  url           = {https://arxiv.org/abs/2106.06452}
}

@inproceedings{zhang2021endtoendurbandrivingimitating,
  title={End-to-end urban driving by imitating a reinforcement learning coach},
  author={Zhang, Zhejun and Liniger, Alexander and Dai, Dengxin and Yu, Fisher and Van Gool, Luc},
  booktitle={Proceedings of the IEEE/CVF international conference on computer vision (ICCV)},
  pages={15222--15232},
  year={2021}
}

@inproceedings{6722531,
  author    = {Solyom, Stefan and Idelchi, Arash and Salamah, Badr Bin},
  booktitle = {2013 IEEE International Conference on Systems, Man, and Cybernetics},
  title     = {Lateral Control of Vehicle Platoons},
  year      = {2013},
  volume    = {},
  number    = {},
  pages     = {4561-4565},
  doi       = {10.1109/SMC.2013.776}
}

@inproceedings{7535437,
  author    = {Bayuwindra, Anggera and Aakre, yvind Løberg and Ploeg, Jeroen and Nijmeijer, Henk},
  booktitle = {2016 IEEE Intelligent Vehicles Symposium (IV)},
  title     = {Combined lateral and longitudinal CACC for a unicycle-type platoon},
  year      = {2016},
  volume    = {},
  number    = {},
  pages     = {527-532}
}

@inproceedings{li2024egostatusneedopenloop,
  title={Is ego status all you need for open-loop end-to-end autonomous driving?},
  author={Li, Zhiqi and Yu, Zhiding and Lan, Shiyi and Li, Jiahan and Kautz, Jan and Lu, Tong and Alvarez, Jose M},
  booktitle={Proceedings of the IEEE/CVF Conference on Computer Vision and Pattern Recognition (CVPR)},
  pages={14864--14873},
  year={2024}
}

@inproceedings{8569947,
  title={A review on cooperative adaptive cruise control (CACC) systems: Architectures, controls, and applications},
  author={Wang, Ziran and Wu, Guoyuan and Barth, Matthew J},
  booktitle={2018 21st International Conference on Intelligent Transportation Systems (ITSC)},
  pages={2884--2891},
  year={2018},
  organization={IEEE}
}

@inproceedings{6427034,
  author    = {Wang, Le Yi and Syed, Ali and Yin, George and Pandya, Abhilash and Zhang, Hongwei},
  booktitle = {2012 IEEE 51st IEEE Conference on Decision and Control (CDC)},
  title     = {Coordinated vehicle platoon control: Weighted and constrained consensus and communication network topologies},
  year      = {2012},
  volume    = {},
  number    = {},
  pages     = {4057-4062}
}

@article{wang2014control,
  title     = {Control of vehicle platoons for highway safety and efficient utility: Consensus with communications and vehicle dynamics},
  author    = {Wang, Le Yi and Syed, Ali and Yin, Gang George and Pandya, Abhilash and Zhang, Hongwei},
  journal   = {Journal of Systems Science and Complexity},
  volume    = {27},
  pages     = {605--631},
  year      = {2014},
  publisher = {Springer}
}

@article{di2014distributed,
  title     = {Distributed consensus strategy for platooning of vehicles in the presence of time-varying heterogeneous communication delays},
  author    = {Di Bernardo, Mario and Salvi, Alessandro and Santini, Stefania},
  journal   = {IEEE Transactions on Intelligent Transportation Systems},
  volume    = {16},
  number    = {1},
  pages     = {102--112},
  year      = {2014},
  publisher = {IEEE}
}

@article{kianfar2012design,
  author    = {Kianfar, R. and Augusto, B. and Ebadighajari, A. and Hakeem, U. and Nilsson, J. and Raza, A. et al.},
  title     = {Design and experimental validation of a cooperative driving system in the grand cooperative driving challenge},
  journal   = {IEEE Transactions on Intelligent Transportation Systems},
  volume    = {13},
  number    = {3},
  pages     = {994--1007},
  year      = {2012}
}

@article{zheng2016distributed,
  title     = {Distributed model predictive control for heterogeneous vehicle platoons under unidirectional topologies},
  author    = {Zheng, Yang and Li, Shengbo Eben and Li, Keqiang and Borrelli, Francesco and Hedrick, J Karl},
  journal   = {IEEE Transactions on Control Systems Technology},
  volume    = {25},
  number    = {3},
  pages     = {899--910},
  year      = {2016},
  publisher = {IEEE}
}

@article{di2015design,
  title     = {Design, analysis, and experimental validation of a distributed protocol for platooning in the presence of time-varying heterogeneous delays},
  author    = {di Bernardo, Mario and Falcone, Paolo and Salvi, Alessandro and Santini, Stefania},
  journal   = {IEEE Transactions on Control Systems Technology},
  volume    = {24},
  number    = {2},
  pages     = {413--427},
  year      = {2015},
  publisher = {IEEE}
}

@article{jia2016platoon,
  title     = {Platoon based cooperative driving model with consideration of realistic inter-vehicle communication},
  author    = {Jia, Dongyao and Ngoduy, Dong},
  journal   = {Transportation Research Part C: Emerging Technologies},
  volume    = {68},
  pages     = {245--264},
  year      = {2016},
  publisher = {Elsevier}
}

@article{geirhos2020shortcut,
  title     = {Shortcut learning in deep neural networks},
  author    = {Geirhos, Robert and Jacobsen, J{\"o}rn-Henrik and Michaelis, Claudio and Zemel, Richard and Brendel, Wieland and Bethge, Matthias and Wichmann, Felix A},
  journal   = {Nature Machine Intelligence},
  volume    = {2},
  number    = {11},
  pages     = {665--673},
  year      = {2020},
  publisher = {Nature Publishing Group UK London}
}

@article{de2019causal,
  title   = {Causal confusion in imitation learning},
  author  = {De Haan, Pim and Jayaraman, Dinesh and Levine, Sergey},
  journal = {Advances in Neural Information Processing Systems},
  volume  = {32},
  year    = {2019}
}

@article{muller2005off,
  title   = {Off-road obstacle avoidance through end-to-end learning},
  author  = {Muller, Urs and Ben, Jan and Cosatto, Eric and Flepp, Beat and Cun, Yann},
  journal = {Advances in Neural Information Processing Systems},
  volume  = {18},
  year    = {2005}
}

@inproceedings{philion2020lift,
  title = {Lift, splat, shoot: Encoding images from arbitrary camera rigs by implicitly unprojecting to 3d},
  author    = {Jonah Philion and Sanja Fidler},
  booktitle = {Proceedings of the European Conference on Computer Vision (ECCV)},
  year      = {2020}
}

@inproceedings{4059340,
  title={A toolbox for easily calibrating omnidirectional cameras},
  author={Scaramuzza, Davide and Martinelli, Agostino and Siegwart, Roland},
  booktitle={2006 IEEE/RSJ International Conference on Intelligent Robots and Systems (IROS)},
  pages={5695--5701},
  year={2006},
  organization={IEEE}
}

@article{9687583,
  title={Visual detection and deep reinforcement learning-based car following and energy management for hybrid electric vehicles},
  author={Tang, Xiaolin and Chen, Jiaxin and Yang, Kai and Toyoda, Mitsuru and Liu, Teng and Hu, Xiaosong},
  journal={IEEE Transactions on Transportation Electrification},
  volume={8},
  number={2},
  pages={2501--2515},
  year={2022},
  publisher={IEEE}
}

@inproceedings{6425157,
  title={Implementation of car-following system using LiDAR detection},
  author={Hsu, Chan Wei and Hsu, Tsung Hua and Chang, Kuang Jen},
  booktitle={2012 12th International Conference on ITS Telecommunications},
  pages={165--169},
  year={2012},
  organization={IEEE}
}

@inproceedings{hu2023planningorientedautonomousdriving,
  title={Planning-oriented autonomous driving},
  author={Hu, Yihan and Yang, Jiazhi and Chen, Li and Li, Keyu and Sima, Chonghao and Zhu, Xizhou and Chai, Siqi and Du, Senyao and Lin, Tianwei and Wang, Wenhai and others},
  booktitle={Proceedings of the IEEE/CVF Conference on Computer Vision and Pattern Recognition (CVPR)},
  pages={17853--17862},
  year={2023}
}

@article{bayliss2009motoring,
  title={Motoring towards 2050—roads and reality background paper No. 9},
  author={Bayliss, D},
  journal={Royal Automobile Club Foundation, Tech. Rep},
  year={2009}
}

@article{maiti2017conceptualization,
  title={A conceptualization of vehicle platoons and platoon operations},
  author={Maiti, Santa and Winter, Stephan and Kulik, Lars},
  journal={Transportation Research Part C: Emerging Technologies},
  volume={80},
  pages={1--19},
  year={2017},
  publisher={Elsevier}
}

@ARTICLE{69979,
   author = {Shladover, S. E. and Desoer, C. A. and Hedrick, J. K. and Tomizuka, M. and Walrand, J. and Zhang, W.-B. et al.},
  journal={IEEE Transactions on Vehicular Technology}, 
  title={Automated vehicle control developments in the PATH program}, 
  year={1991},
  volume={40},
  number={1},
  pages={114-130}
}

@inproceedings{fritz1999longitudinal,
  title={Longitudinal and lateral control of heavy duty trucks for automated vehicle following in mixed traffic: experimental results from the CHAUFFEUR project},
  author={Fritz, Hans},
  booktitle={Proceedings of the 1999 IEEE International Conference on Control Applications},
  volume={2},
  pages={1348--1352},
  year={1999},
  organization={IEEE}
}

@inproceedings{bergenhem2010challenges,
  title={Challenges of platooning on public motorways},
  author={Bergenhem, Carl and Huang, Qihui and Benmimoun, Ahmed and Robinson, Tom},
  booktitle={17th World Congress on Intelligent Transport Systems},
  pages={1--12},
  year={2010}
}

@article{bergenhem2012vehicle,
  title={Vehicle-to-vehicle communication for a platooning system},
  author={Bergenhem, Carl and Hedin, Erik and Skarin, Daniel},
  journal={Procedia-Social and Behavioral Sciences},
  volume={48},
  pages={1222--1233},
  year={2012},
  publisher={Elsevier}
}

@inproceedings{yi2023lidar,
  title={A LiDAR-assisted smart car-following framework for autonomous vehicles},
  author={Yi, Xianyong and Ghazzai, Hakim and Massoud, Yehia},
  booktitle={2023 IEEE International Symposium on Circuits and Systems (ISCAS)},
  pages={1--5},
  year={2023},
  organization={IEEE}
}
\bibliographystyle{IEEEtran}

\clearpage
\vspace{250pt}
\begin{IEEEbiography}
[{\includegraphics[width=1in,height=1.25in,clip,keepaspectratio]{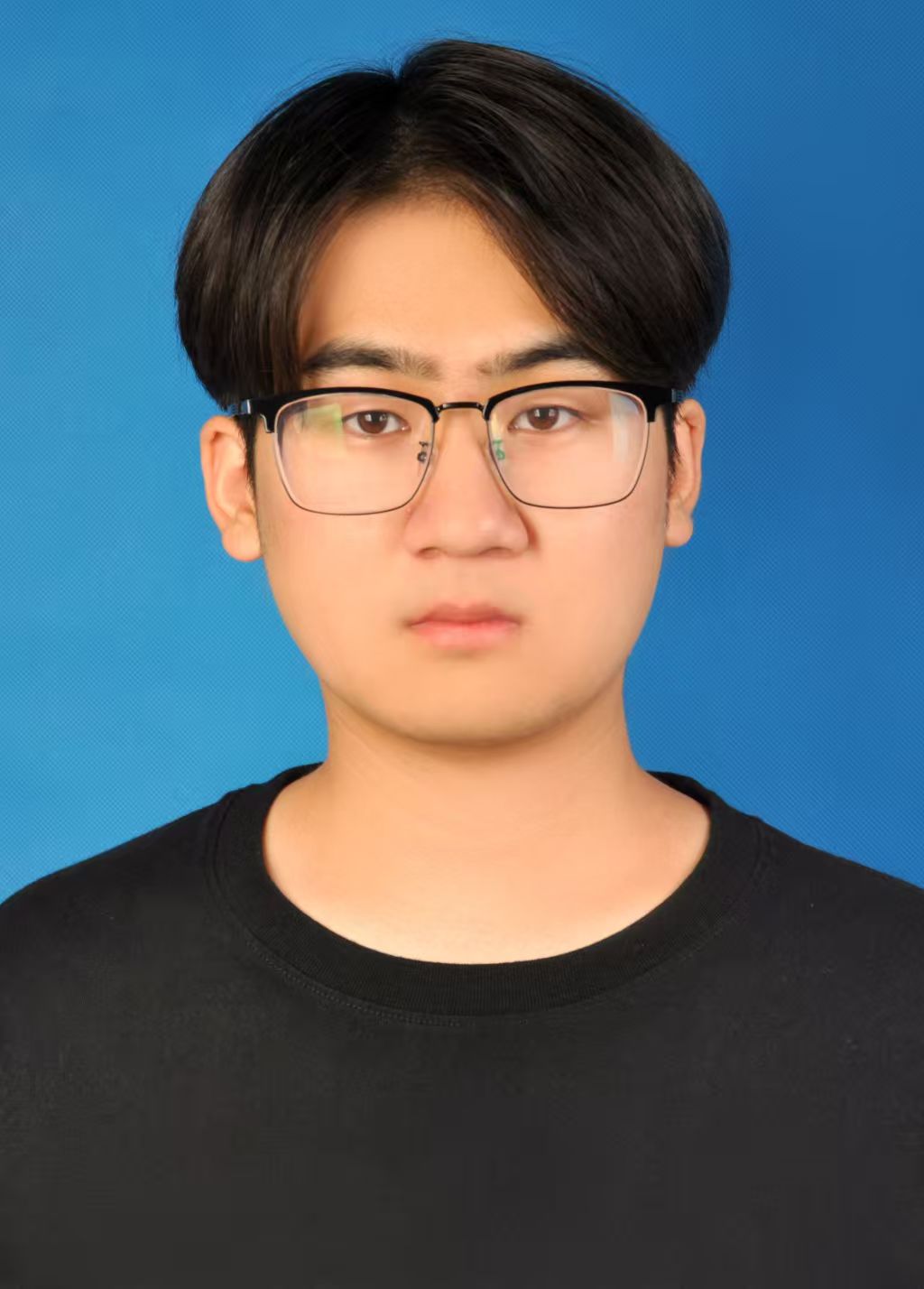}}]{Jiale Zhang} 
    received a Bachelor's degree in Flight Vehicle Control and Information Engineering from Northwestern Polytechnical University, Xi’an, China, in 2023. He is currently pursuing a Ph.D. degree in Control Science and Engineering at Shanghai Jiao Tong University. His main research interests include end-to-end autonomous driving, decision-making algorithms, and reinforcement learning for intelligent vehicles.
\end{IEEEbiography}

\begin{IEEEbiography}
[{\includegraphics[width=1in,height=1.25in,clip,keepaspectratio]{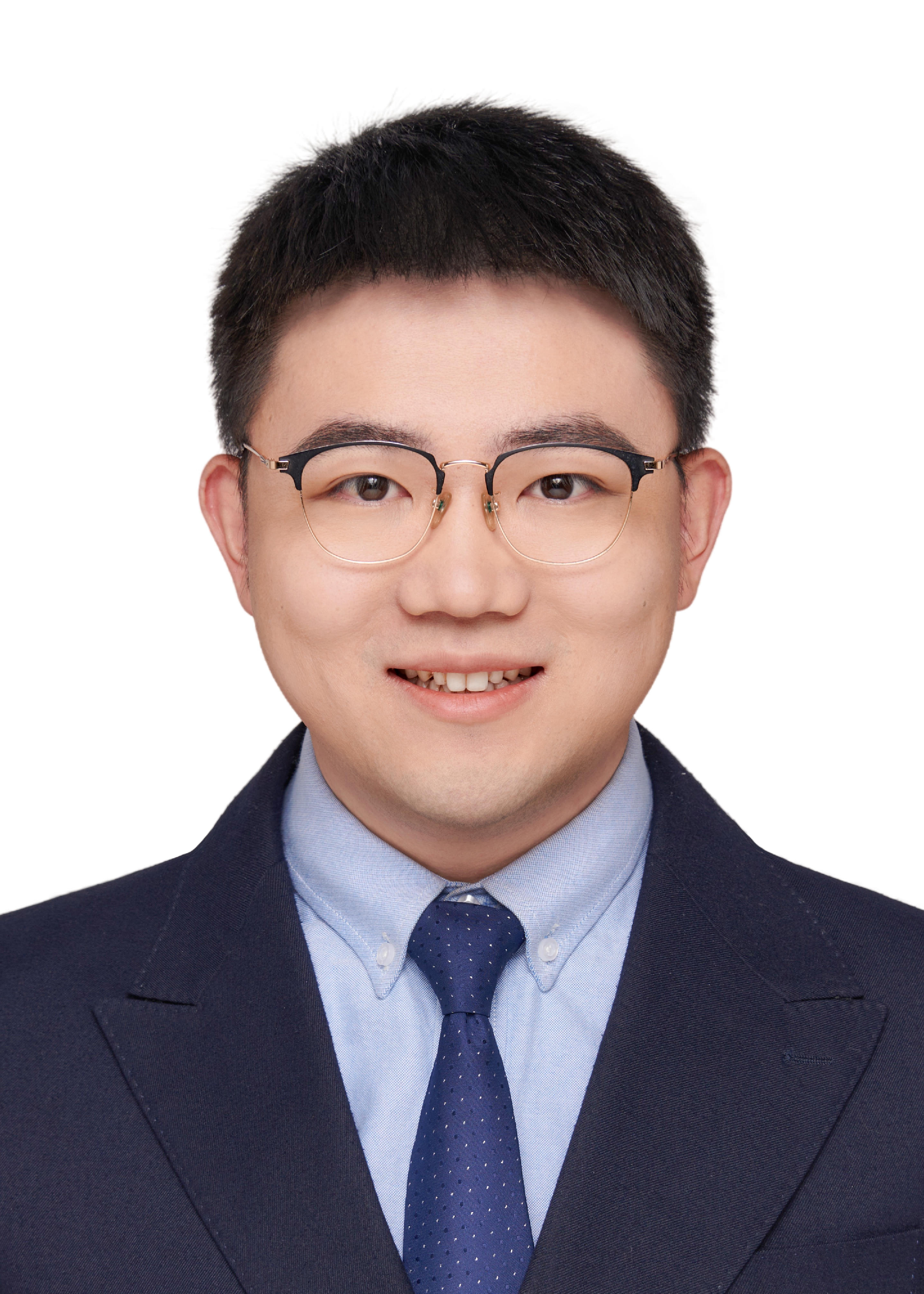}}]{Yeqiang Qian} 
    received his Ph.D. degree in control science and engineering from Shanghai Jiao Tong University, Shanghai, China, in 2020. He is currently the tenure track associate professor with the Department of Automation at Shanghai Jiao Tong University. His main research interests include computer vision, pattern recognition, machine learning and their applications.
\end{IEEEbiography}

\begin{IEEEbiography}
[{\includegraphics[width=1in,height=1.25in,clip,keepaspectratio]{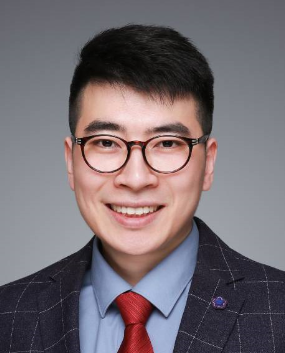}}]{Tong qin} 
    received the B.Eng degree in control science and engineering from the Zhejiang University, China, in 2015, and the Ph.D. degree in the Department of Electronic and Computer Engineering, the Hong Kong University of Science and Technology, HongKong, in 2019. He worked as a staff research scientist in the department of Advanced Driving Solution, Huawei from 2019 to 2023. He is currently working as an associate professor in Global Institute of Future Technology, Shanghai Jiao Tong University. His research interests include SLAM, NeRF, machine learning, and autonomous driving.
\end{IEEEbiography}

\begin{IEEEbiography}
[{\includegraphics[width=1in,height=1.25in,clip,keepaspectratio]{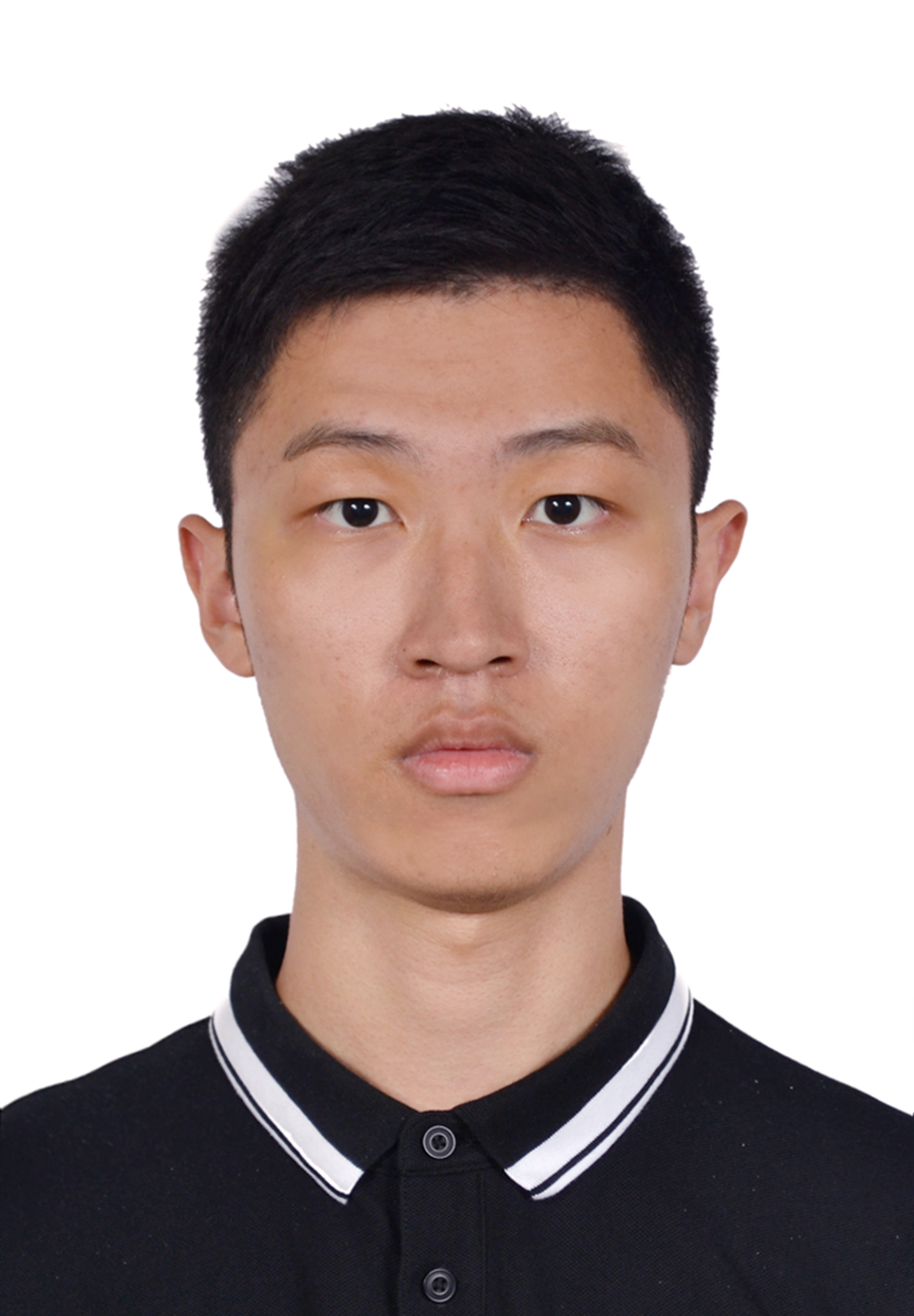}}]{Mingyang Jiang} 
    received a Bachelor's degree in engineering from Shanghai Jiao Tong University, Shanghai, China, in 2023. He is working towards a Master's degree in Control Science and Engineering from Shanghai Jiao Tong University. His main research interests are end-to-end planning, driving decision-making, and reinforcement learning for autonomous vehicles.
\end{IEEEbiography}

\begin{IEEEbiography}[{\includegraphics[width=1in,height=1.25in,clip,keepaspectratio]{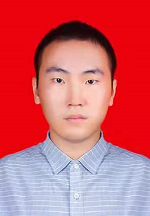}}]{Siyuan Chen}
    received a Bachelor's degree in engineering from Shanghai Jiao Tong University, Shanghai, China, in 2023. He is working towards a Master's degree in Control Science and Engineering from Shanghai Jiao Tong University. His main research interests are planning and control, V2X system for autonomous vehicles.
\end{IEEEbiography}

\begin{IEEEbiography}[{\includegraphics[width=1in,height=1.25in,clip,keepaspectratio]{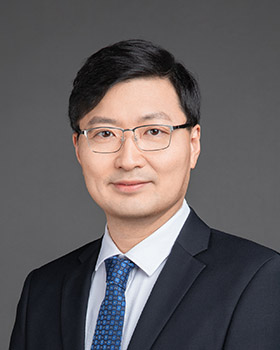}}]{Ming YANG}
    received his Master’s and Ph.D. degrees from Tsinghua University, Beijing, China, in 1999 and 2003, respectively. Presently, he holds the position of Distinguished Professor at Shanghai Jiao Tong University, also serving as the Director of the Innovation Center of Intelligent Connected Vehicles. Dr. Yang has been engaged in the research of intelligent vehicles for more than 25 years.
\end{IEEEbiography}

\end{document}